\PassOptionsToPackage{unicode}{hyperref}
\PassOptionsToPackage{hyphens}{url}
\documentclass[
]{article}
\usepackage{amsmath,amssymb}
\usepackage{iftex}
\ifPDFTeX
  \usepackage[T1]{fontenc}
  \usepackage[utf8]{inputenc}
  \usepackage{textcomp} 
\else 
  \usepackage{unicode-math} 
  \defaultfontfeatures{Scale=MatchLowercase}
  \defaultfontfeatures[\rmfamily]{Ligatures=TeX,Scale=1}
\fi
\usepackage{lmodern}
\ifPDFTeX\else
\fi
\IfFileExists{upquote.sty}{\usepackage{upquote}}{}
\IfFileExists{microtype.sty}{
  \usepackage[]{microtype}
  \UseMicrotypeSet[protrusion]{basicmath} 
}{}
\makeatletter
\@ifundefined{KOMAClassName}{
  \IfFileExists{parskip.sty}{%
    \usepackage{parskip}
  }{
    \setlength{\parindent}{0pt}
    \setlength{\parskip}{6pt plus 2pt minus 1pt}}
}{
  \KOMAoptions{parskip=half}}
\makeatother
\usepackage{xcolor}
\usepackage{longtable,booktabs,array}
\usepackage{calc} 
\usepackage{etoolbox}
\makeatletter
\patchcmd\longtable{\par}{\if@noskipsec\mbox{}\fi\par}{}{}
\makeatother
\IfFileExists{footnotehyper.sty}{\usepackage{footnotehyper}}{\usepackage{footnote}}
\makesavenoteenv{longtable}
\usepackage{graphicx}
\makeatletter
\def\maxwidth{\ifdim\Gin@nat@width>\linewidth\linewidth\else\Gin@nat@width\fi}
\def\maxheight{\ifdim\Gin@nat@height>\textheight\textheight\else\Gin@nat@height\fi}
\makeatother
\setkeys{Gin}{width=\maxwidth,height=\maxheight,keepaspectratio}
\makeatletter
\def\fps@figure{htbp}
\makeatother
\providecommand{\tightlist}{%
  \setlength{\itemsep}{0pt}\setlength{\parskip}{0pt}}
\NewDocumentCommand\citeproctext{}{}

\makeatletter
 \let\@cite@ofmt\@firstofone
 \def\@biblabel#1{}
 \def\@cite#1#2{{#1\if@tempswa , #2\fi}}
\makeatother
\newlength{\cslhangindent}
\newlength{\csllabelwidth}
\newenvironment{CSLReferences}[2] 
 {\begin{list}{}{%
  \setlength{\itemindent}{0pt}
  \setlength{\leftmargin}{0pt}
  \setlength{\parsep}{0pt}
  \ifodd #1
   \setlength{\leftmargin}{\cslhangindent}
   \setlength{\itemindent}{-1\cslhangindent}
  \fi
  \setlength{\itemsep}{#2\baselineskip}}}
 {\end{list}}
\usepackage{calc}

\newcommand{\CSLLeftMargin}[1]{\parbox[t]{\csllabelwidth}{\strut#1\strut}}
\newcommand{\CSLRightInline}[1]{\parbox[t]{\linewidth - \csllabelwidth}{\strut#1\strut}}

\usepackage{amsmath, amssymb, amsfonts}
\usepackage{booktabs}
\usepackage{graphicx}
\usepackage{geometry}
\usepackage{caption}
\makeatletter
\@ifpackageloaded{subfig}{}{\usepackage{subfig}}
\@ifpackageloaded{caption}{}{\usepackage{caption}}
\AtBeginDocument{%

}
\AtBeginDocument{%

}
\newcounter{pandoccrossref@subfigures@footnote@counter}
{\end{figure}%
\addtocounter{footnote}{-\value{pandoccrossref@subfigures@footnote@counter}}
\@for\f:=\global@pandoccrossref@subfigures@footnotes\do{\stepcounter{footnote}\footnotetext{\f}}%
\gdef\global@pandoccrossref@subfigures@footnotes{}}
\@ifpackageloaded{float}{}{\usepackage{float}}
\floatstyle{ruled}
\@ifundefined{c@chapter}{\newfloat{codelisting}{h}{lop}}{\newfloat{codelisting}{h}{lop}[chapter]}
\floatname{codelisting}{Listing}

\makeatother
\ifLuaTeX
  \usepackage{selnolig}  
\fi
\usepackage{bookmark}
\IfFileExists{xurl.sty}{\usepackage{xurl}}{} 
\hypersetup{
  pdftitle={Epistemic Compression: The Case for Deliberate Ignorance in High-Stakes AI},
  pdfauthor={Steffen Lukas},
  pdfkeywords={Artificial Intelligence, Model Robustness, Foundation
Models, Epistemic Compression, Clinical AI, Rate Reduction, White-Box
Deep Learning},
  hidelinks,
  pdfcreator={LaTeX via pandoc}}

\title{Epistemic Compression: The Case for Deliberate Ignorance in
High-Stakes AI}
\usepackage{etoolbox}
\makeatletter
\providecommand{\subtitle}[1]{
  \apptocmd{\@title}{\par {\large #1 \par}}{}{}
}
\makeatother
\subtitle{Why Complexity Backfires in High-Stakes AI}
\author{Steffen Lukas}
\date{March 2026}

\begin{document}
\maketitle

\begin{center}
Charité – Universitätsmedizin Berlin, Germany\\
\texttt{steffen.lukas@charite.de}
\end{center}
\begin{abstract}

Foundation models excel in stable environments, yet often fail where
reliability matters most---medicine, finance, policy. This
\emph{Fidelity Paradox} is not just a data problem; it is structural. In
domains where rules change over time, extra model capacity amplifies
noise rather than capturing signal. We introduce \emph{Epistemic
Compression}: the principle that robustness emerges from matching model
complexity to the ``shelf life'' of the data, not from scaling
parameters. Unlike classical regularization, which penalizes weights
post-hoc, Epistemic Compression enforces parsimony \emph{through
architecture}---the model structure itself is designed to reduce
overfitting by making it architecturally costly to represent variance
that exceeds the evidence in the data. We operationalize this with a
\emph{Regime Index} that separates \emph{Shifting Regime} (unstable,
data-poor---simplicity wins) from \emph{Stable Regime} (invariant,
data-rich---complexity viable). In an exploratory synthesis of 15
high-stakes domains, this index was concordant with the empirically
superior modeling strategy in 86.7\% of cases (13/15). High-stakes AI
demands a shift from scaling for its own sake to \emph{principled
parsimony}.

\end{abstract}

\section{The Fidelity Paradox}\label{the-fidelity-paradox}

A persistent puzzle runs through quantitative science. Massive neural
networks solve protein folding\textsuperscript{1}, yet struggle with
clinical prognosis and financial forecasting\textsuperscript{2--5}. We
call this the \emph{Fidelity Paradox}: the problem is not that these
models cannot fit the data---they fit it too well. They memorize noise.
In medicine and finance, the real challenge is knowing what to
\emph{ignore}. Simple portfolios beat sophisticated optimization in
noisy markets\textsuperscript{6}. Basic clinical scores outperform deep
learning under distribution shift\textsuperscript{7,8}. Yet the
prevailing research direction continues to prioritize
scale\textsuperscript{9,10}.

The COVID-19 pandemic provided a natural experiment confirming this
paradox. Systematic reviews identified hundreds of diagnostic AI tools
developed for COVID-19 detection; the vast majority were judged at high
risk of bias, and few demonstrated robust external
validation\textsuperscript{9,11}. These models learned
shortcuts---hospital tags, patient positioning markers, even the word
``portable'' on radiographs---rather than pathological
features\textsuperscript{12}. The problem was not insufficient data or
compute; it was that high-capacity models are \emph{designed} to find
patterns, and in the Shifting Regime, the loudest patterns are often
spurious.

We propose \emph{Epistemic Compression}: Ockham's Razor as an algorithm.
The insight is old---``the map is not the
territory''\textsuperscript{13}---but the application is urgent. In
unstable environments, \emph{deliberate ignorance} is not a bug but a
feature. Models that leave out complexity on purpose are often the only
ones that survive when the world changes.

\section{The Mechanics of Robustness}\label{the-mechanics-of-robustness}

We use \emph{high-capacity models} to mean overparameterized
architectures like deep neural networks, with \emph{foundation models}
(pretrained on massive data) as the leading example. But scale and
complexity are not the same thing. We define complexity as
\emph{Effective Dimensionality (\(\mathbf{D}_{\text{eff}}\))}: how many
independent numbers you need to describe the model's core
behavior\textsuperscript{14}. A billion-parameter model can be
effectively simple if its structure forces it to ignore most of its
capacity. Robustness, then, is not about having fewer parameters---it is
about having the \emph{right constraints} (Figure 1).

\begin{figure}
\centering
\includegraphics{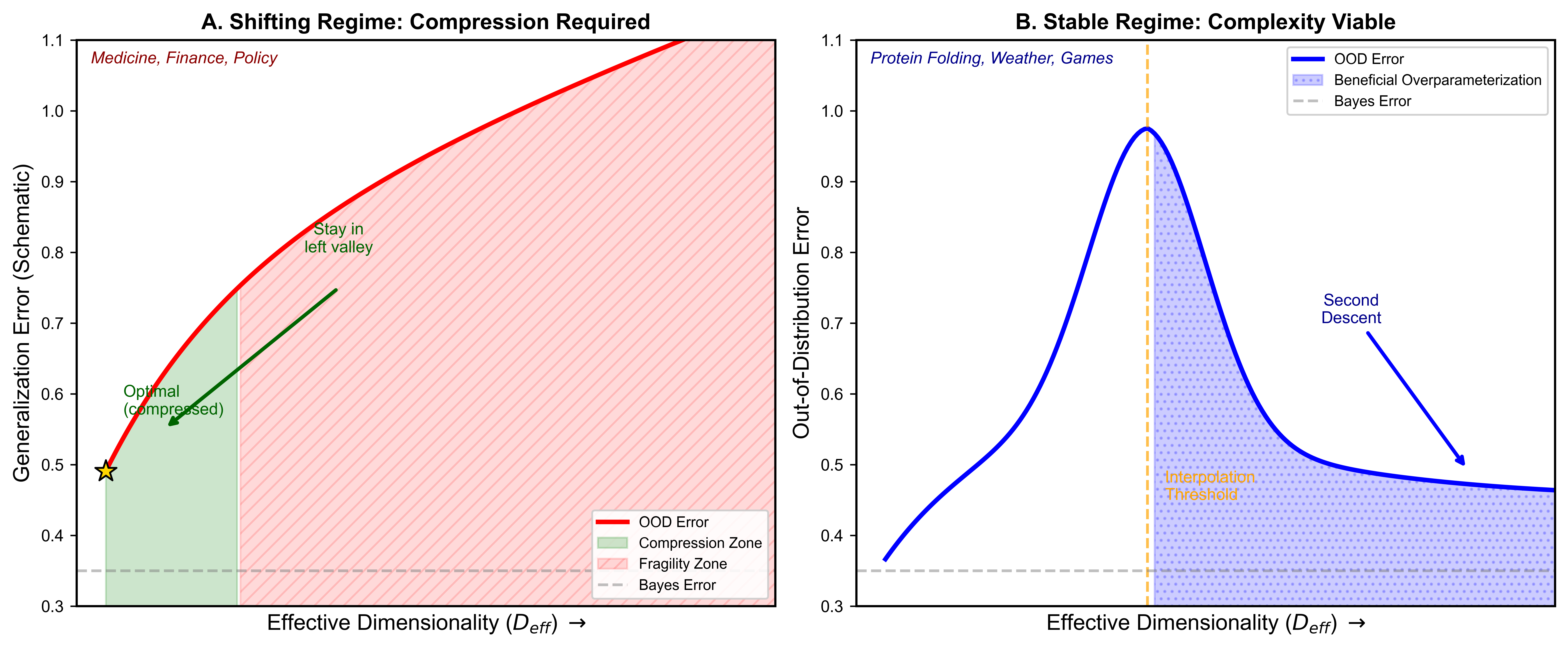}
\caption{The geometry of generalization by regime. \emph{a,} In the
Shifting Regime (low signal stability), OOD error (red curve) rises
monotonically after an optimal compression point (\(D^*\)), creating a
``Fragility Zone'' (red shading) where excess capacity fits spurious
correlations. The optimal strategy is ``staying in the left valley''
(green shading). \emph{b,} In the Stable Regime (high signal stability),
the ``double descent'' phenomenon occurs\textsuperscript{15}, where
massive overparameterization (blue shading) reduces error by capturing
fine-grained invariant structure. Note that double descent applies to
in-distribution performance on a fixed data distribution; its benefits
do not extend to the distribution-shifted setting that characterizes the
Shifting Regime.}\label{fig:geometry}
\end{figure}

\emph{Epistemic Compression differs fundamentally from classical
regularization.} Techniques like L1/L2 penalties or dropout impose
constraints \emph{after} the architecture is specified---they shrink
weights but do not change what the model can represent. Epistemic
Compression operates \emph{through} architecture: the model structure
itself limits what can be learned. A pharmacokinetic compartment model
cannot overfit drug timing artifacts because mass-action kinetics are
hard-coded into its equations. A CRATE layer cannot memorize noise
because its compression objective geometrically prevents it. This is the
difference between a leash (regularization) and a fence (architecture).

\subsection{The Two Mechanisms of
Robustness}\label{the-two-mechanisms-of-robustness}

\emph{Type A Compression (Structural Isolation)} finds the true
mechanistic backbone and throws everything else away. Newtonian
mechanics (\(F=ma\)) is Type A: it ignores relativity because, for
everyday speeds, relativity does not matter\textsuperscript{16,17}. Type
A models work by maximizing \emph{Rate Reduction}---finding the simplest
representation that still separates the classes\textsuperscript{18}.
Problems like protein folding are \emph{computationally hard} (huge
search spaces) but \emph{epistemically simple} (governed by stable
physics), so Type A compression can succeed if you have enough compute.

\emph{Type B Compression (Defensive Compression)} is a protective move.
It works not by finding truth, but by admitting that in a noisy,
shifting world, trying to estimate the ``true'' relationship is a losing
game. The 1/N portfolio ignores correlations between stocks because
those correlations change too fast to estimate reliably. This logic
connects to core ideas in machine learning: the \emph{Information
Bottleneck}\textsuperscript{19,20}, \emph{Structural Risk
Minimization}\textsuperscript{21}, and \emph{ecological
rationality}\textsuperscript{22,23}. When the environment is noisy,
strong assumptions are often the only defense against fitting garbage.
This distinction parallels Kahneman's System 1/System 2
dichotomy\textsuperscript{24} and recent proposals for ``System 2 Deep
Learning''\textsuperscript{25}---the recognition that fast
pattern-matching (System 1) fails where deliberate reasoning (System 2)
is required.

The danger zone is \emph{high-capacity models applied to noisy, shifting
data} (e.g., Transformers on electronic health records). These models
sit in a ``fragility zone'' (Figure 1): powerful enough to memorize
training noise, but without the built-in constraints to tell noise from
signal. The fragility zone lies at intermediate complexity---the model
has outgrown the safety of simplicity but has not acquired the causal
structure that would make complexity safe. The transition is gradual
(Figure 2b), with steepness depending on the signal-to-noise ratio. We
quantify this shortfall as the \emph{Viability Gap} (see below): the
distance between how much data you have and how much you need for robust
generalization.

\begin{figure}
\centering
\includegraphics{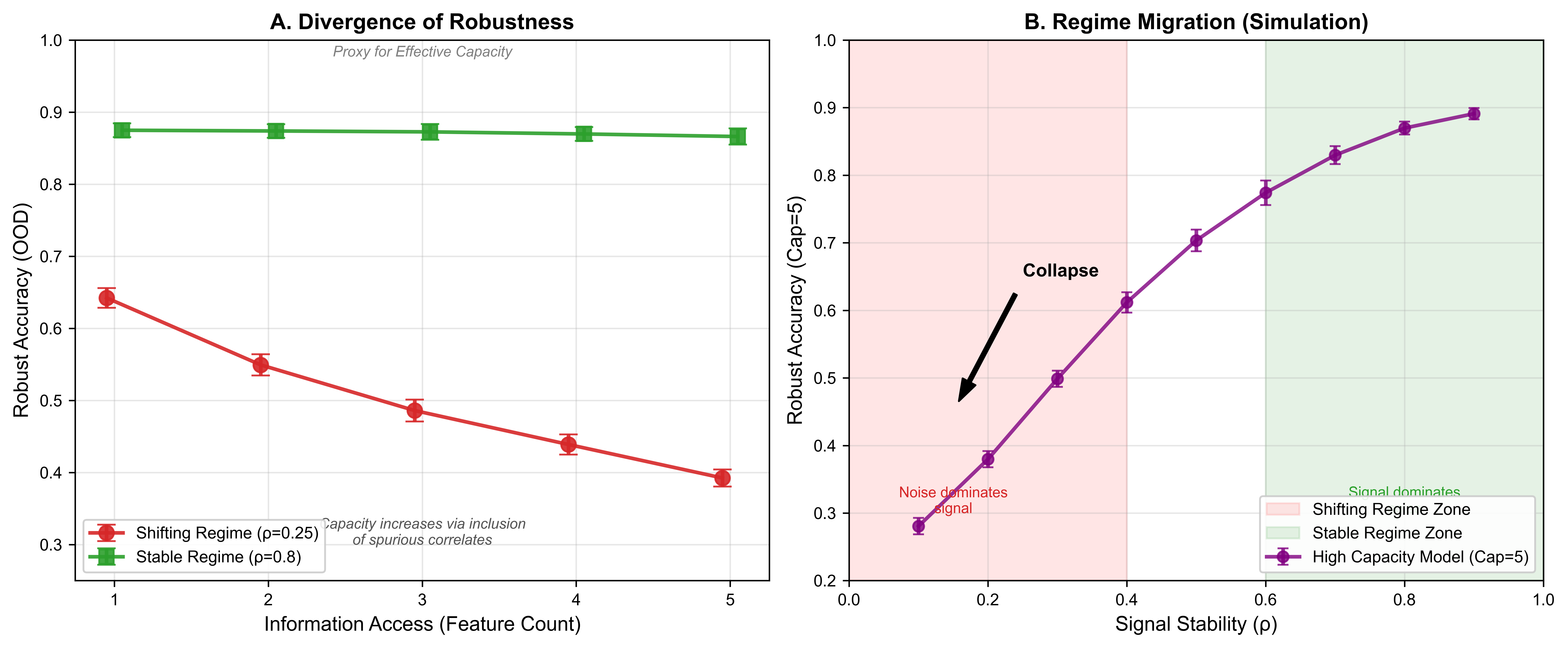}
\caption{Empirical simulation of the robustness--capacity tradeoff.
\emph{a,} Robust accuracy (minimum of in-distribution and OOD accuracy)
versus model capacity (number of accessible features). In Shifting
Regime environments (red line, \(\rho=0.25\)), increasing
capacity---defined here as the inclusion of spurious
covariates---degrades robustness (\(\downarrow\)) as the model exploits
non-stationary signals. In the Stable Regime (green line, \(\rho=0.8\)),
the invariant signal dominates, rendering capacity increases benign.
\emph{b,} Regime migration: at fixed high capacity, robustness improves
sigmoidally with signal stability (\(\rho\)), illustrating the phase
transition between regimes. The inflection point suggests a critical
threshold where the dominant signal transitions from noise to invariant
structure. Error bars denote s.d. over \(n=20\)
repetitions.}\label{fig:capacity}
\end{figure}

\section{The Two Regimes and the Regime
Index}\label{the-two-regimes-and-the-regime-index}

We propose a practical \emph{Regime Index (RI)} to diagnose when
complexity is a liability.

\subsection{The Shifting Regime}\label{the-shifting-regime}

This regime is characterized by \emph{Non-stationarity} (rules change
over time)\textsuperscript{26}, \emph{Data Poverty} (low
samples-to-dimension ratio)\textsuperscript{27},
\emph{Underspecification}\textsuperscript{28}, and \emph{Weak Ground
Truth} (proxy labels). Here, complexity is \emph{maladaptive}. Fitting
fine structure improves in-distribution accuracy but degrades
robustness. Most high-stakes societal problems---medicine, finance,
policy---are the Shifting Regime.

\subsection{The Stable Regime}\label{the-stable-regime}

This regime has \emph{stable rules}, \emph{abundant data}, and
\emph{objective ground truth}. Here, foundation models genuinely shine.
\emph{AlphaFold}\textsuperscript{1} and
\emph{AlphaZero}\textsuperscript{29} succeed because physics and game
rules do not change. With enough data from a fixed distribution, massive
models can interpolate the underlying structure with stunning precision.
Cancer imaging foundation models\textsuperscript{30} similarly thrive
when biological signals are stable and standardized.

\subsection{The Regime Index}\label{the-regime-index}

We operationalize this distinction into a five-indicator diagnostic
score with three tiers (Figure 3): \emph{3--5 = Shifting} (Compression
Mandatory), \emph{1.5--2.5 = Borderline} (Compression Superiority Test
required), \emph{0--1 = Stable} (Complexity Viable). The indicators
include temporal stability, context invariance, data-to-complexity ratio
(\(N/D_{\text{eff}}\)), ground truth quality, and causal prior strength
(Box 1). The data-to-complexity threshold (\(N/D_{\text{eff}}\)
\textless{} 100 for the Shifting criterion) aligns with classical
statistical power requirements: stable coefficient estimation requires
approximately 10--20 samples per predictor\textsuperscript{31,32}, and
we apply a 5--10x safety margin for covariate shift and model selection
variance. In an exploratory synthesis of 15 high-stakes domains, the
full Regime Index was concordant with the empirically superior modeling
strategy in 86.7\% of cases (13/15)---providing initial triangulation
for the framework rather than a formally validated meta-analytic result.

\begin{figure}
\centering
\includegraphics{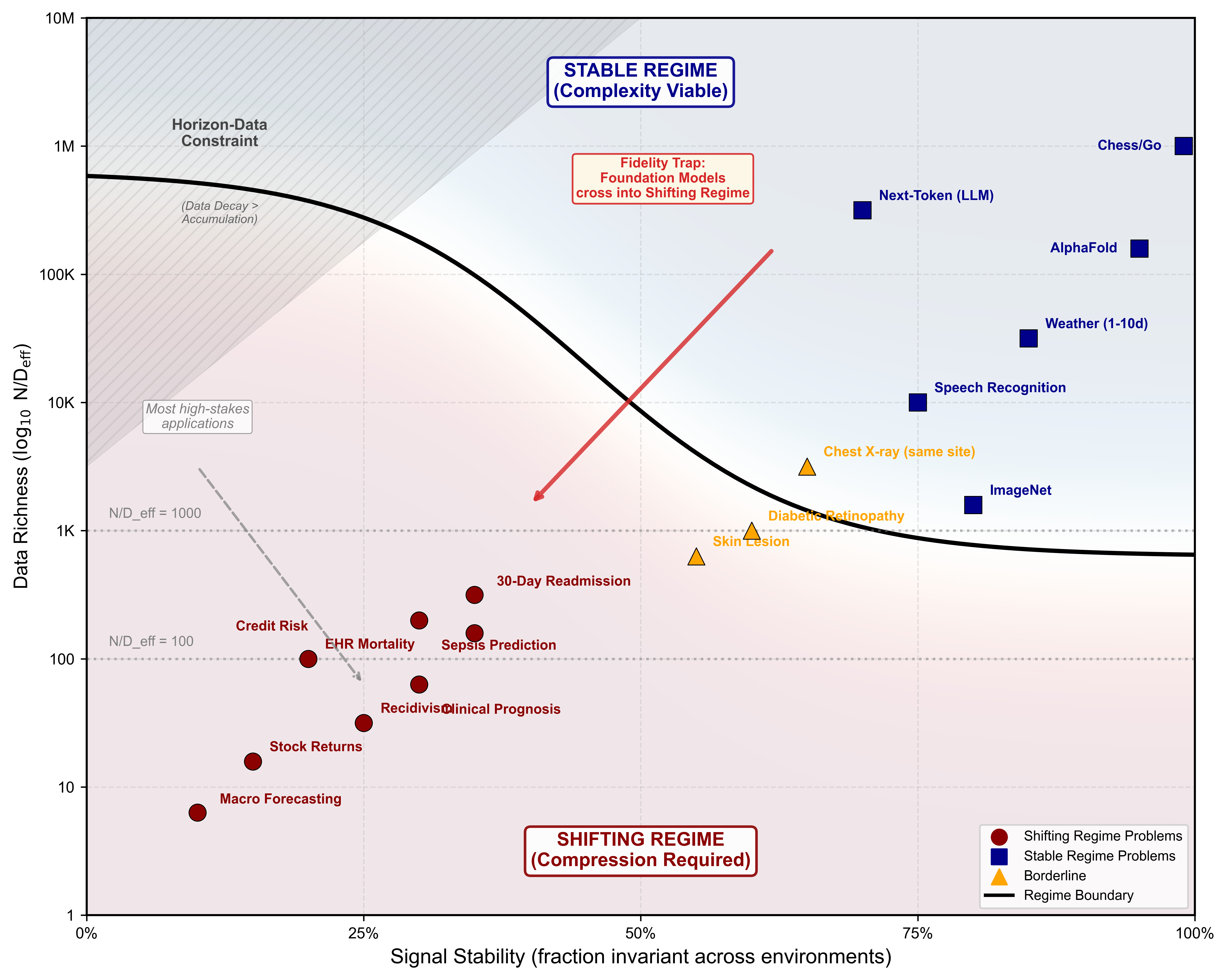}
\caption{The regime phase diagram. A diagnostic map for high-stakes
model selection. The boundary curve (black line) is a conceptual
frontier delineating the minimum data richness
(\(\log_{10} N/D_{\text{eff}}\)) required for robust generalization at a
given level of signal stability (derived in Supplementary Note 5). The
shaded region (top-left) indicates the ``Horizon-Data Constraint,''
where non-stationarity prevents the accumulation of sufficient data to
train high-capacity models---data half-life creates a physical ceiling
on achievable sample size. Domain positions represent illustrative
schematic placements drawn from and beyond the 15-domain formal
synthesis; all positions are approximate and not authoritative
coordinates. Domains are grouped by regime: red circles (Shifting), blue
squares (Stable), orange triangles (Borderline/mixed). For the complete
formal scoring of the 15 synthesis domains, see Supplementary Table 1.
High-stakes societal applications (red circles) are shown clustered in
the Shifting Regime based on this benchmark evidence. Foundation models
trained on massive data (blue squares) occupy the Stable Regime but risk
the ``Fidelity Trap'' (red arrow) when applied to non-stationary
Shifting Regime tasks.}\label{fig:phase_diagram}
\end{figure}

The Regime Index (RI) operationalizes the \emph{No Free Lunch}
theorem\textsuperscript{33} and modern statistical learning
theory\textsuperscript{31} for the era of foundation models. As detailed
in \emph{Box 1} (see end of manuscript), it scores domains on five
binary indicators: temporal stability, context invariance,
data-to-complexity ratio, ground truth quality, and causal prior
strength. Scores of 3--5 indicate the Shifting Regime (Compression
Mandatory); scores of 1.5--2.5 are Borderline (requiring the Compression
Superiority Test); scores of 0--1 indicate the Stable Regime (Complexity
Viable). Critically, \emph{Temporal Instability is a necessary condition
for Stable classification}: any domain exhibiting documented
non-stationarity must be treated as at minimum Borderline regardless of
total score, since temporal drift overrides data richness as a
robustness constraint.

\subsection{The Viability Gap}\label{the-viability-gap}

We introduce the \emph{Viability Gap} (\(\mathcal{V}\)) as a
quantitative diagnostic for regime classification:

\[\mathcal{V} = \log_{10}(N/D_{\text{eff}}) - \mathcal{B}(\rho)\]

where \(\mathcal{B}(\rho)\) is an illustrative boundary representing the
approximate minimum data richness for robust generalization at signal
stability \(\rho\) (Supplementary Note 5; note: \(\mathcal{B}(\rho)\) is
a conceptual diagnostic geometry, not a calibrated estimator). Domains
with \(\mathcal{V} < 0\) operate in \emph{structural deficit}: no amount
of hyperparameter tuning can compensate for the fundamental mismatch
between model capacity and data utility. This formalizes the intuition
that some problems are not ``hard'' in the computational sense but
``impossible'' in the epistemological sense---the data simply cannot
support the conclusions we ask models to draw.

\section{Empirical Evidence: The Regime
Hypothesis}\label{empirical-evidence-the-regime-hypothesis}

We illustrate the regime hypothesis through three real-world domains
characterized by distribution shift. Throughout, we report: (i)
\emph{training AUROC} (performance on the training distribution), (ii)
\emph{OOD AUROC} (performance on shifted test distributions), and (iii)
\emph{robust AUROC} (minimum across temporal/demographic splits). The
``winner'' in the Shifting Regime is determined by \emph{stability}
(smaller \(\Delta\)), not absolute OOD performance.

\emph{1. Financial Shift (LendingClub):} In a re-analysis of 1.35
million public loan records (75,088 training loans, 668,651 far-term
evaluation loans), a Feature-Tokenization Transformer
(FT-Transformer\textsuperscript{34})---a leading deep learning model for
tabular data---achieved the highest training AUROC (0.720) yet suffered
the largest degradation under economic shift (2011--2016, fully matured
loans; \(\Delta\)AUROC +0.039). A Gradient Boosting Machine showed
intermediate degradation (\(\Delta\)AUROC +0.020). A simple expert
logistic regression (8 FICO/DTI-based features) actually \emph{improved}
as the economic cycle matured (\(\Delta\)AUROC \(-0.020\))---an
anti-fragility pattern absent in higher-capacity models. Across the
capacity ladder, the highest-capacity models were the most fragile, with
the FT-Transformer degrading nearly 2x more than GBM despite similar
training AUROC.

\emph{2. Clinical Deterioration (MIMIC-IV):} In ICU mortality prediction
using MIMIC-IV v3.1 (68,546 ICU stays, 2008--2019), a high-capacity MLP
(with standard L2 regularization) achieved near-perfect training AUROC
(0.954) but degraded substantially under temporal shift (0.954 \(\to\)
0.760, \(\Delta\) = 0.194; worst-case 0.740). A simple logistic
regression with expert-selected physiological features showed only a
modest temporary dip during the ICD-10 transition (0.733 in Env A, 0.716
in Env B, 0.739 in Env C; robust AUROC 0.716, \(\Delta\) = -0.006 from
Env A to Env C). The compression advantage---0.20 AUROC points less
degradation---demonstrates that in non-stationary clinical environments,
deliberate model simplicity preserves predictive validity.

\emph{3. Meta-Synthesis (15 Domains):} To test the Regime Index, we
synthesized evidence from 15 published high-stakes domains
(Supplementary Note 1). To reduce selection bias, we restricted the
analysis to domains satisfying three pre-specified inclusion criteria:
high-stakes decision authority, temporal shift benchmarks, and published
complexity comparisons. We acknowledge that this necessarily excludes
domains where complexity wins unambiguously (e.g., language modeling),
which may be underrepresented in shift-evaluation literature. Despite
this limitation, the synthesis is striking: the Regime Index was
concordant with the superior modeling strategy in 13 of 15 cases (Table
1). We emphasize that this synthesis is exploratory and
hypothesis-generating, not a formal meta-analysis; its value lies in
triangulating the framework across diverse domains, not in providing
definitive validation.

\begin{longtable}[]{@{}
  >{\raggedright\arraybackslash}p{(\columnwidth - 8\tabcolsep) * \real{0.2000}}
  >{\raggedright\arraybackslash}p{(\columnwidth - 8\tabcolsep) * \real{0.2000}}
  >{\raggedright\arraybackslash}p{(\columnwidth - 8\tabcolsep) * \real{0.2000}}
  >{\raggedright\arraybackslash}p{(\columnwidth - 8\tabcolsep) * \real{0.2000}}
  >{\raggedright\arraybackslash}p{(\columnwidth - 8\tabcolsep) * \real{0.2000}}@{}}
\caption{\label{tbl:retrospective}\emph{Summary of the 15-Domain
Exploratory Synthesis.} \emph{Concordance} denotes the proportion of
domains where the Regime Index was concordant with the empirically
superior model class. \emph{Statistical Analysis}: Overall concordance
is 86.7\% (13/15; 95\% CI {[}62.1\%--96.3\%{]}, Wilson score interval;
two-sided Binomial Test, \(H_0: p=0.5\), \(p < 0.01\)). This synthesis
is illustrative and hypothesis-generating, not a formal meta-analysis;
the 15 domains were selected to span the regime spectrum rather than to
constitute a representative sample of all AI applications. Full domain
scoring and references are provided in Supplementary
Information.}\tabularnewline
\toprule\noalign{}
\begin{minipage}[b]{\linewidth}\raggedright
Regime
\end{minipage} & \begin{minipage}[b]{\linewidth}\raggedright
Examples
\end{minipage} & \begin{minipage}[b]{\linewidth}\raggedright
Mean RI
\end{minipage} & \begin{minipage}[b]{\linewidth}\raggedright
Winning Strategy
\end{minipage} & \begin{minipage}[b]{\linewidth}\raggedright
Concordance
\end{minipage} \\
\midrule\noalign{}
\endfirsthead
\toprule\noalign{}
\begin{minipage}[b]{\linewidth}\raggedright
Regime
\end{minipage} & \begin{minipage}[b]{\linewidth}\raggedright
Examples
\end{minipage} & \begin{minipage}[b]{\linewidth}\raggedright
Mean RI
\end{minipage} & \begin{minipage}[b]{\linewidth}\raggedright
Winning Strategy
\end{minipage} & \begin{minipage}[b]{\linewidth}\raggedright
Concordance
\end{minipage} \\
\midrule\noalign{}
\endhead
\bottomrule\noalign{}
\endlastfoot
\emph{Shifting Regime} & Medicine, Finance, Policy & 3.0 &
\emph{Compression} (Simple Models) & \emph{80\%} (8/10) \\
\emph{Stable Regime} & Physics, Games, Vision & 0.2 & \emph{Complexity}
(Deep Learning) & \emph{100\%} (5/5) \\
\end{longtable}

The two discordant cases---psychiatric readmission and income
prediction---offer instructive lessons. In psychiatric readmission, the
complex model ``won'' on the held-out metric, but \emph{post-hoc}
failure analysis revealed it had learned administrative shortcuts (e.g.,
medication timestamps correlated with nursing shift changes) rather than
clinical deterioration. In income prediction (UCI
Adult\textsuperscript{35}), complex models showed less degradation under
demographic shift than simple models, though this domain's 1994-era data
may limit generalizability to modern contexts. These exceptions
reinforce our claim: in the Shifting Regime, in-distribution accuracy is
an unreliable proxy for robustness, and careful domain-specific
evaluation remains essential.

\section{The Limits of Scale}\label{the-limits-of-scale}

Scaling is the correct solvent for complexity in \emph{Stable Regime}.
In domains governed by stable, objective laws---such as protein folding
(AlphaFold) or weather forecasting---massive overparameterization
succeeds because it interpolates a fixed ground truth. Here, scale is
virtuous.

\subsection{The Horizon-Data Constraint and Data
Half-Life}\label{the-horizon-data-constraint-and-data-half-life}

A critical observation emerges when plotting these domains: they adhere
to a diagonal frontier (Fig. 3). In non-stationary regimes, data
possesses a finite ``half-life.'' We define \emph{Data Half-Life}
(\(\tau_{1/2}\)) as the time window after which model performance
degrades by 50\% of its initial advantage over a naive baseline.
Illustrative estimates: for ICU mortality prediction, \(\tau_{1/2}\) is
on the order of years (driven by ICD coding transitions, protocol
updates, and EMR migrations); for consumer credit scoring,
\(\tau_{1/2}\) spans economic cycles (1--3 years); for fraud detection,
\(\tau_{1/2}\) can be less than a year due to adversarial adaptation.

This creates a \emph{Horizon-Data Constraint} that is structural, not
merely resource-limited. Let \(\dot{N}\) denote the rate of stable data
accumulation and \(N^* = c \cdot D_{\text{eff}}\) the minimum sample
size required for robust generalization. In the Shifting Regime,
\(\tau_{1/2}\) bounds the viable accumulation window, giving the
constraint:

\[N_{\text{viable}} = \tau_{1/2} \cdot \dot{N} \ll N^*\]

For ICU mortality (MIMIC-IV): \(\tau_{1/2} \approx 3\) years,
\(\dot{N} \approx 7{,}000\) stable stays/year,
\(N_{\text{viable}} \approx 21{,}000\), while
\(N^* \approx 100 \times D_{\text{eff}} \geq 1{,}200\) --- at face value
sufficient. But the effective \(D_{\text{eff}}\) under distribution
shift far exceeds the expert-feature count, and the clock resets with
each guideline change. For fraud detection (\(\tau_{1/2} < 1\) year),
the constraint is always binding. Consequently, for Shifting Regime
problems, ``scaling up'' is not merely a resource challenge but a
\emph{chronological impossibility}: the required stability window does
not exist.

\subsection{Can Technical Advances Overcome the Shifting
Regime?}\label{can-technical-advances-overcome-the-shifting-regime}

One might argue that better pre-training, domain adaptation, or
synthetic data could push a domain from the Shifting Regime into the
Stable Regime. We think this conflates \emph{fixable} problems with
\emph{unfixable} ones. Better sensors can reduce measurement noise.
Federated learning can broaden sampling. But nothing can eliminate
\emph{ontological non-stationarity}---the fact that clinical guidelines
change, economies shift, and people adapt to interventions. This is not
a data quality issue; it is a property of the world. Synthetic data
works beautifully in the Stable Regime (AlphaZero's self-play), but in
the Shifting Regime it risks amplifying the very artifacts you are
trying to escape\textsuperscript{36}. The Horizon-Data Constraint is
\emph{structural}, not \emph{technical}: you cannot stabilize an
inherently moving target.

\emph{Continual learning does not resolve the Shifting Regime}---it
merely delays collapse. Each model update requires labeled data from the
\emph{new} distribution, which is expensive and often unavailable until
after deployment failures have occurred. The model is always chasing a
moving target. Moreover, frequent retraining introduces its own risks:
model ``churn'' (where predictions flip between versions) undermines
clinical trust, and regulatory frameworks struggle to approve
continuously updating systems. In adversarial domains like fraud
detection, the retraining cycle itself becomes a
vulnerability---fraudsters can probe the model, trigger an update, and
exploit the transition period.

A related objection: what about \emph{causal discovery} methods that
identify invariant structure? If we could learn the causal graph, could
we convert Type B problems into Type A? In principle, yes---this is an
exciting frontier. In practice, causal discovery in high-noise,
high-dimensional settings faces severe identifiability constraints and
sample complexity limits that currently preclude reliable application in
most Shifting Regime domains\textsuperscript{37}. We view this as an
important avenue for future work, but not a near-term solution.

\subsection{Scaling Works for Interpolation, Not
Extrapolation}\label{scaling-works-for-interpolation-not-extrapolation}

In closed systems with infinite stationary data, scaling dissolves
complexity. But scaling laws describe \emph{interpolation} on fixed
distributions. They say nothing about \emph{extrapolation} to shifted
ones\textsuperscript{8,38,39}. Empirical audits show that while scaling
improves perplexity, it often fails to improve downstream
reliability\textsuperscript{40}. Algorithmic approaches---Invariant Risk
Minimization\textsuperscript{38}, distributionally robust
optimization\textsuperscript{41}, anchor
regression\textsuperscript{42}---address shift by modifying training
objectives. Epistemic Compression is complementary but prior: it
addresses architecture rather than objective, asking what
\emph{structure} a model must have to be inherently shift-resistant.
When a model must extrapolate, the ``certainty-scope trade-off'' kicks
in\textsuperscript{43}: a model that tries to cover more ground will be
less certain about any given point. Current foundation models lack the
self-verification mechanisms needed to stay
grounded\textsuperscript{44,45}.

Recent phenomena like ``grokking''\textsuperscript{46}---where
generalization emerges after extended training---are likely limited to
\emph{Stable Regime (Closed Systems)} where a fixed ground truth exists.
In the Shifting Regime, where distributions drift, relying on emergent
generalization is structurally equivalent to waiting for the model to
memorize transient noise.

This invites a reconsideration of Sutton's ``Bitter
Lesson''\textsuperscript{47,48}---the claim that brute-force computation
beats clever human design. We argue this is a \emph{Stable Regime
theorem}. In the Shifting Regime, we observe the opposite: \emph{the
Efficient Lesson}. Simple methods that encode domain knowledge
outperform raw compute. When signal-to-noise is low, more computation
just means more noise amplification. A massive neural network
\emph{will} find patterns in stock market data---but they will be
spurious.

The Efficient Lesson deserves emphasis: in the Shifting Regime, the
``human bottleneck'' is not a limitation---it is a \emph{filter}. Domain
experts act as information-theoretic sieves, rejecting spurious
correlations before they can poison the model. The physician who refuses
to include ``time of admission'' as a predictor is not being Luddite;
she is enforcing causal priors that no amount of data can learn. The
Bitter Lesson assumed infinite stationary data; in finite, shifting
worlds, clever priors beat brute force. This is why building ``machines
that learn and think like people''\textsuperscript{49}---with strong
inductive biases and compositional structure---remains essential for the
Shifting Regime deployment.

\section{Why High-Capacity Models Struggle in the Shifting
Regime}\label{why-high-capacity-models-struggle-in-the-shifting-regime}

The failure of high-capacity models in the Shifting Regime is not
random---it is mechanistic, driven by four compounding pathologies.

First: the \emph{Recursion Trap}\textsuperscript{36}. Recursive
self-improvement works in the Stable Regime (AlphaZero) because
synthetic data interpolates a stable ground truth. In the Shifting
Regime, it backfires. As models train on their own outputs, the
distribution collapses toward the mode, smoothing away the rare ``tail''
events that matter most for risk assessment.

Second: \emph{Structural Mismatch}. Foundation models learn
\emph{universal} patterns (grammar, edges) from internet-scale data. But
high-stakes decisions often hinge on \emph{local, transient}
signals---hospital-specific workflows, regional economic quirks. These
require \emph{abductive reasoning}\textsuperscript{50} and \emph{tacit
knowledge}\textsuperscript{51} that pretraining cannot provide. When
fine-tuned, models suffer \emph{Inductive
Interference}\textsuperscript{10}: they import strong but spurious
correlations from pretraining. A systematic survey of over 80 foundation
models trained on electronic medical records found that most are
evaluated on narrow benchmarks that obscure this structural mismatch,
and that current assessments fail to demonstrate consistent value in
real-world healthcare deployment\textsuperscript{52}.

Third: the \emph{Attention-to-Noise Mechanism}. In high-noise
environments, attention heads become ``confounder-seeking missiles,''
latching onto shortcuts like hospital tags in COVID-19
radiographs\textsuperscript{12,53}. As illustrated in
\emph{Supplementary Figure 1}, a Transformer can attend perfectly to a
spurious artifact (e.g., an ``L'' marker), minimizing training loss
while destroying robust accuracy. A simpler model, lacking this
resolution, effectively ``blurs'' over the artifact. (Note: Attention
maps in Supp. Fig 1 are schematic simulations.)

Fourth: the \emph{Fluency-Fidelity Gap}\textsuperscript{54}. Models
optimized for human preference (RLHF) achieve \emph{Rhetorical
Alignment}---they sound plausible---rather than \emph{Epistemic
Alignment}---knowing when to say ``I don't know.'' The result is
confident hallucinations that create real harm\textsuperscript{54,55}.
This gap is starkly visible in medical applications: GPT-4 achieves
impressive scores on medical licensing exams, yet exhibits systematic
failure modes---hallucination, miscalibration, and brittleness to
context variation---when applied to real clinical reasoning tasks
requiring integration of patient-specific
context\textsuperscript{55--57}. The model can recite textbook knowledge
(the Stable Regime) but struggles to navigate the uncertainty inherent
in individual patient care (the Shifting Regime). Recent audits in
pathology confirm this fragility\textsuperscript{58}. Techniques like
RLHF\textsuperscript{59} and LoRA\textsuperscript{60} improve
instruction-following and efficiency, but they address surface
alignment, not the deeper structural mismatch.

\section{The Path Forward: Principled
Complexity}\label{the-path-forward-principled-complexity}

We are not advocating a retreat to logistic regression. We propose
\emph{Principled Complexity}: using architectural constraints to enforce
parsimony, rather than relying on data volume to smooth over variance.
Crucially, white-box architectures are not defined by low parameter
count---a CRATE layer can have millions of parameters, while a
pharmacokinetic compartment model has four---but by \emph{hard
structural priors} that prevent fitting what should be ignored. The
constraint is architectural, not quantitative. This manifests in
\emph{Mathematically Interpretable Architectures}---models derived from
first principles (conservation laws, symmetry groups, compression
objectives) rather than heuristic search. Examples include
\emph{Kolmogorov-Arnold Networks}\textsuperscript{61}, \emph{Liquid
Neural Networks}\textsuperscript{62}, \emph{Geometric Deep
Learning}\textsuperscript{63}, and \emph{TabPFN}\textsuperscript{64},
which scales \emph{inductive priors} rather than parameters.

A prime example is the \emph{CRATE} architecture\textsuperscript{65}.
Unlike standard Transformers, CRATE optimizes a white-box compression
objective (\emph{Rate Reduction}). As shown in Figure 4 and
Supplementary Note 4, these architectures are \emph{regime-adaptive}:
they scale expressivity when data is abundant (Stable Regime) but
default to sparse, linear-like constraints when data is scarce (Shifting
Regime). This structural bias toward compression is designed to reduce
the overfitting that plagues black-box models, by making it
architecturally costly to represent variance that exceeds the
information content of the data.

\begin{figure}
\centering
\includegraphics{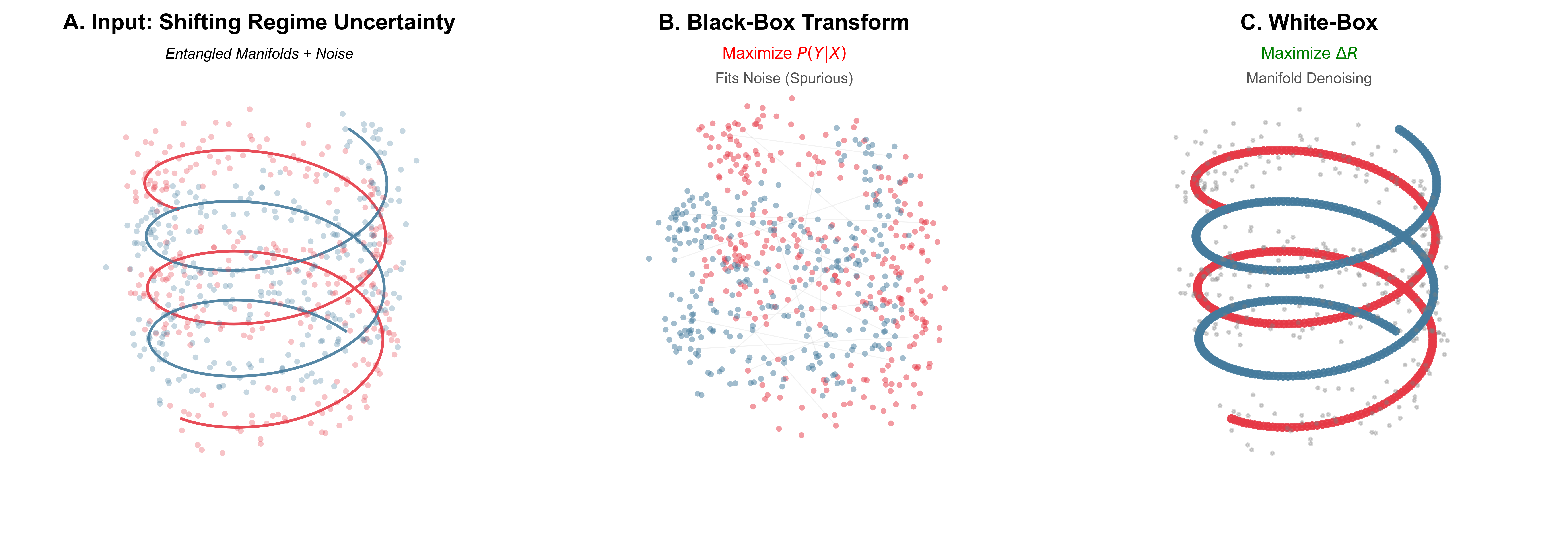}
\caption{Geometric conceptualization of epistemic compression. \emph{a,}
The Shifting Regime data characterized by low-dimensional manifolds
entangled with high-variance noise. \emph{b,} Black-box behavior
(schematic): likelihood maximization unconstrains the feature space,
fitting specific noise instances to separate classes, resulting in
high-energy, brittle representations. \emph{c,} White-box behavior
(compression-based architecture): optimization of rate reduction
(\(\Delta R\)) acts as a geometric sieve, collapsing noise onto the
underlying low-dimensional manifold and orthogonalizing class subspaces,
thereby recovering the invariant causal structure.}\label{fig:whitebox}
\end{figure}

We propose a workflow of \emph{Structured Distillation}: 1.
\emph{Hypothesis Generation (the Stable Regime)}: Use an LLM to scan
literature and suggest candidate causal features. 2. \emph{Epistemic
Filtering (the Shifting Regime)}: A human expert reviews candidates,
filtering out spurious correlates. 3. \emph{Compressed Modeling (the
Shifting Regime)}: Train a white-box model using only the validated
features.

This aligns with recent calls for cascaded human-AI
systems\textsuperscript{44,66}. A concrete example is
\emph{Model-Informed Precision Dosing}: standard deep learning models
often overfit to spurious signals (e.g., the timing of blood draws),
while \emph{pharmacokinetic compartment models} enforce mass-action
kinetics, compressing patient state into \textasciitilde4 invariant
parameters. This structural constraint acts as an epistemic
firewall---predictions remain physiologically valid even for novel
dosing regimens.

As deployment safeguard, we recommend an \emph{Epistemic Firewall}: a
lightweight, interpretable ``Gatekeeper Model'' running in parallel with
any high-capacity system. If the complex model's output diverges
significantly from the Firewall's baseline, the system triggers a
\emph{failsafe halt}.

\section{Implications and
Predictions}\label{implications-and-predictions}

The choice between compression and complexity is not just technical---it
is ethical. Black-box models create an \emph{accountability gap}. They
exacerbate \emph{systemic risk} through \emph{Algorithmic
Monoculture}\textsuperscript{67,68}: if a foundation model has a blind
spot, it gets replicated across institutions. High-capacity models are
also expensive to recalibrate when distributions shift; simple models
can be updated with minimal data, aligning with the \emph{precautionary
principle}. High-stakes decisions demand models that are inherently
interpretable---not just ``explainable'' after the
fact\textsuperscript{69}. This offers \emph{safety through transparency}
and mitigates \emph{algorithm aversion}\textsuperscript{70}---the rapid
erosion of human trust after seeing an algorithm err.

A limitation of this work is that while we synthesized evidence from 15
domains, this Perspective does not include a formal risk-of-bias
assessment (e.g., ROBINS-I) of the cited studies, which varies by domain
maturity. Additionally, because shift-evaluated simple-vs-complex
comparisons are not uniformly reported across machine learning, our
synthesis is necessarily conditional on the subset of high-stakes
domains with published OOD evaluations. Importantly, this conditioning
likely introduces a \emph{reversal bias} that works against our
conclusion: complex models are more likely to have published OOD
evaluations precisely when those evaluations are favorable. That even in
this publication-biased subset, simple models win 80\% of the Shifting
Regime cases strengthens, not weakens, the framework's empirical
support. Future work should systematically audit these benchmarks for
label leakage and expand the synthesis using pre-registered
meta-analytic methods\textsuperscript{71}.

For practitioners, we recommend the \emph{Compression Superiority Test
(CST)}: 1. \emph{Train Baseline}: Fit a transparent, low-capacity model
(Type B) using expert-selected features. 2. \emph{Train Challenger}: Fit
the high-capacity model (e.g., Transformer) on the same data. 3.
\emph{Evaluate \(\Delta\)}: Compute performance on a \emph{shifted} test
set. 4. \emph{Decision}: Only adopt complexity if it yields a
significant improvement (\(> \delta\)) OOD. Otherwise, default to
simplicity.

We anchor this framework in three falsifiable hypotheses for the
2026--2031 period:

\begin{enumerate}
\def\labelenumi{\arabic{enumi}.}
\item
  \emph{Clinical Risk (by 2029)}: In prospective, multi-center
  validation trials, black-box foundation models will often fail to
  reliably outperform validated simple scores (e.g., APACHE-II, qSOFA,
  LACE) by a clinically significant margin (\(\geq\) 0.05 AUROC) in
  common prediction tasks (mortality, readmission, sepsis).
\item
  \emph{Regulatory Correction (by 2030)}: We predict that multiple
  currently FDA-cleared or CE-marked black-box clinical AI tools will
  face withdrawal, restriction, or requirements for human-in-the-loop
  safeguards due to demonstrated cross-site performance degradation.
\item
  \emph{The White-Box Mandate (by 2031)}: We anticipate that regulatory
  bodies (FDA, EMA, or equivalent) will move toward guidance favoring
  \emph{constructive interpretability}---architectures derived from
  domain principles---over post-hoc explainability for AI systems making
  autonomous high-stakes decisions (risk class III or
  equivalent)\textsuperscript{11,69,72}.
\end{enumerate}

These predictions are falsifiable: if by 2031 foundation models
consistently outperform simple scores in prospective trials, our
framework is wrong. We invite the field to test it.

\section{Conclusion}\label{conclusion}

We are entering an era of industrial-scale scientific modeling. The
danger is that we have built engines of complexity (Stable Regime tools)
and are applying them indiscriminately to domains of uncertainty
(Shifting Regime). True intelligence is not just the ability to learn
patterns---it is the discipline to ignore them. Epistemic compression is
the architectural enforcement of that discipline.

\emph{In high-stakes AI, less is more: the models that survive are the
ones that knew what to forget.}

Future progress will favor approaches that optimize for parsimony:
finding the simplest representation that supports robust
decision-making. This is not a retreat from deep learning. It is its
evolution toward a rigorous \emph{Architectural Science}---one that
recognizes the epistemic limits of learning from data and builds those
limits into the structure of our models.

\section{Methods}\label{methods}

\emph{LendingClub analysis.} 1,345,350 fully matured loan records
(issued 2011--2016, final status resolved by 2018Q4) were retrieved from
the public LendingClub dataset\textsuperscript{73}. Three
non-overlapping temporal environments were pre-specified: Env A
(2011--2012, n = 75,088, training period), Env B (2013--2014, n =
357,907, near-term shift), and Env C (2015--2016, n = 668,651, far-term
shift). All models were trained exclusively on Env A and evaluated
without re-fitting on Env B and Env C. Six model classes spanning a
capacity ladder were compared: (1) logistic regression on 8
expert-selected features (loan amount, interest rate, income, DTI, FICO,
open accounts, revolving utilization, employment length); (2) logistic
regression on 20 extended features; (3) gradient boosting (200
estimators, max depth 3); (4) shallow MLP (1 hidden layer, 32 units);
(5) deep MLP (2 hidden layers, 128x64 units); and (6)
Feature-Tokenization Transformer (FT-Transformer\textsuperscript{34};
\(d_\text{token}=64\), 3 Transformer layers, 4 attention heads, FFN
width 256, AdamW optimiser, lr \(3\times10^{-4}\), 50 epochs, batch size
2048). All features were standardised (mean 0, unit variance) using
training-set statistics. Robust AUROC was defined as
\(\min(\mathrm{AUROC}_A, \mathrm{AUROC}_B, \mathrm{AUROC}_C)\);
degradation \(\Delta = \mathrm{AUROC}_A - \mathrm{AUROC}_C\).

\emph{MIMIC-IV analysis.} 68,546 adult ICU stays (2008--2019) from
MIMIC-IV v3.1 on PhysioNet\textsuperscript{74,75} were divided into
three pre-specified temporal environments: Env A (2008--2014, training),
Env B (2015--2016, ICD-10 transition period), and Env C (2017--2019,
post-transition evaluation). MIMIC-IV is a deidentified electronic
health record resource made available to credentialed researchers who
complete human-subject protections training and sign a data use
agreement; the Beth Israel Deaconess Medical Center Institutional Review
Board granted a waiver of informed consent and approved sharing of the
resource\textsuperscript{75}. Logistic regression (9 expert
physiological features: age, sex, heart rate, systolic blood pressure,
respiratory rate, temperature, SpO\(_2\), GCS score, ICU length of stay)
and MLP (all available numeric features with L2 regularization) were
trained on Env A and evaluated without re-fitting on Env B and Env C.
Outcome: in-hospital mortality at ICU discharge.

\emph{15-domain synthesis.} Studies were identified through a structured
search and selected by three pre-specified criteria: (1) high-stakes
decision context with real-world consequences; (2) temporal or
cross-site distribution-shift evaluation; and (3) published comparison
of at least two model-complexity tiers. The Regime Index was scored
independently using the Box 1 rubric; any score differences were
resolved by discussion. Statistical analysis used a two-sided binomial
test (\(H_0\): \(p\) = 0.5) with 95\% confidence intervals by the Wilson
score method. Full domain scores and citations are provided in
Supplementary Table 1.

\emph{Reproducibility.} Figures 1--4 are reproducible from the analysis
procedures described in the manuscript and supplementary materials.
Reproducing the MIMIC-IV analyses additionally requires credentialed
PhysioNet access.

\section{Data availability}\label{data-availability}

The datasets analyzed in this study are publicly available but require
separate download:

\begin{itemize}
\tightlist
\item
  \emph{LendingClub}: Loan data available at
  \href{https://www.kaggle.com/datasets/wordsforthewise/lending-club}{kaggle.com/datasets/wordsforthewise/lending-club}
  (free Kaggle account required).
\item
  \emph{UCI Adult Income}: Available at
  \href{https://archive.ics.uci.edu/ml/datasets/adult}{archive.ics.uci.edu/ml/datasets/adult}
  (direct download); distributed by the UCI Machine Learning Repository
  under CC BY 4.0\textsuperscript{35}.
\item
  \emph{MIMIC-IV}: Available on PhysioNet
  (\href{https://physionet.org/content/mimiciv/}{physionet.org/content/mimiciv});
  access is limited to credentialed users who complete CITI Data or
  Specimens Only Research training and sign the PhysioNet Credentialed
  Health Data Use Agreement\textsuperscript{74--76}.
\item
  \emph{15-Domain Meta-Synthesis}: Relies on published results cited in
  the Supplementary Information; no additional data download required.
\end{itemize}

\clearpage

\section{Boxes}\label{boxes}

\emph{Box 1: The Regime Index Scorecard}

The Regime Index classifies problems into \emph{Shifting Regime}
(High-stakes, data-poor; Compression Mandatory) or \emph{Stable Regime}
(Stable, data-rich; Complexity Viable)\textsuperscript{77}. While
\(D_{\text{eff}}\) is formally the intrinsic dimension of the causal
manifold, practitioners can estimate it using three complementary
approaches (detailed in Supplementary Note 6): \emph{(1) Domain
Priors}---the number of expert-validated variables in clinical
guidelines or regulatory standards (e.g., APACHE-II uses 12 variables,
suggesting \(D_{\text{eff}} \leq 12\) for ICU severity); \emph{(2)
Statistical Estimation}---intrinsic dimensionality estimators such as
PCA eigenvalue decay or two-nearest-neighbor
methods\textsuperscript{78}; \emph{(3) Learning Curves}---the sample
size at which validation performance plateaus, which approximates
\(N^* \approx 10 \times D_{\text{eff}}\) under standard learning
theory\textsuperscript{31}. The threshold \(N/D_{\text{eff}}\)
\textless{} 100 aligns with classical statistical power requirements for
stable coefficient estimation in regression, with a safety margin for
distribution shift.

\begin{longtable}[]{@{}
  >{\raggedright\arraybackslash}p{(\columnwidth - 4\tabcolsep) * \real{0.3333}}
  >{\raggedright\arraybackslash}p{(\columnwidth - 4\tabcolsep) * \real{0.3333}}
  >{\raggedright\arraybackslash}p{(\columnwidth - 4\tabcolsep) * \real{0.3333}}@{}}
\toprule\noalign{}
\begin{minipage}[b]{\linewidth}\raggedright
Indicator
\end{minipage} & \begin{minipage}[b]{\linewidth}\raggedright
Shifting Regime Criteria (Score +1)
\end{minipage} & \begin{minipage}[b]{\linewidth}\raggedright
Stable Regime Criteria (Score 0)
\end{minipage} \\
\midrule\noalign{}
\endhead
\bottomrule\noalign{}
\endlastfoot
\emph{1. Temporal Stability} & \emph{Unstable}: Relationships degrade
(\(\Delta\)AUROC \textgreater{} 0.05) over 2--5 years. (e.g., Clinical
practice, Finance) & \emph{Stable}: Mechanisms are invariant over
decades. (e.g., Physics, Protein folding) \\
\emph{2. Context Invariance} & \emph{Low}: Models fail to transfer
across institutions without retraining. & \emph{High}: Data generation
is standardized globally. (e.g., Mass spec, Go rules) \\
\emph{3. Data-to-Complexity} & \emph{Poverty}:
\(N/D_{\text{eff}} < 100\). (Transition Zone: 100--1000). &
\emph{Abundance}: \(N/D_{\text{eff}} > 1000\). Data saturates signal
complexity. \\
\emph{4. Ground Truth} & \emph{Subjective/Proxy}: Labels are noisy
proxies (e.g., ``readmission'') or low-agreement (\(\kappa < 0.8\)). &
\emph{Objective}: Labels are unambiguous (e.g., ``Win/Loss'', ``Crystal
structure''). \\
\emph{5. Causal Priors} & \emph{Weak}: Causal graph is unknown,
contested, or unencodable. & \emph{Strong}: Invariant physical laws can
be encoded as constraints. \\
\end{longtable}

\emph{Scoring \& Recommendation}: * \emph{Score 3--5 (Shifting Regime)}:
Prioritize \emph{Epistemic Compression}. Use Type B (Defensive) or Type
A (Structural Isolation) approaches. High-capacity black boxes are
\emph{deleterious}. * \emph{Score 1.5--2.5 (Borderline)}: Conduct a
\emph{Compression Superiority Test} (CST; see Protocol). Do not deploy
complex models without direct shift-robustness evidence. * \emph{Score
0--1 (Stable Regime)}: Prioritize \emph{Complexity}. Scale model
capacity and data. * \emph{Temporal Instability Gate}: Stable
classification requires Indicator 1 (Temporal Stability) = 0. If
Indicator 1 scores 1, the domain is Borderline at minimum regardless of
the other indicators. Temporal stability is a \emph{necessary condition}
for the Stable Regime: a data-rich but temporally non-stationary domain
cannot be classified as Stable.

\emph{Scoring note}: Each indicator is scored 0 (Stable criterion met)
or 1 (Shifting criterion met). Half-scores (0.5) may be assigned for
indicators that are clearly present but attenuated; these must be
explicitly documented in the domain scoring record. For Indicator 3, if
the available \(D_{\text{eff}}\) estimation methods straddle a threshold
boundary, assign 0.5 unless independent domain evidence justifies a more
decisive score. Total scores are summed and interpreted directly: 0--1 =
Stable, 1.5--2.5 = Borderline, 3--5 = Shifting.

\clearpage
\setcounter{figure}{0}
\renewcommand{\thefigure}{S\arabic{figure}}
\renewcommand{\theHfigure}{S\arabic{figure}}
\setcounter{table}{0}
\renewcommand{\thetable}{S\arabic{table}}
\renewcommand{\theHtable}{S\arabic{table}}

\section{Supplementary Information}\label{supplementary-information}

\subsection{Supplementary Note 1: Methodology for the 15-Domain
Meta-Synthesis}\label{supplementary-note-1-methodology-for-the-15-domain-meta-synthesis}

\subsubsection{1. Domain Selection
Criteria}\label{domain-selection-criteria}

We selected 15 predictive domains from the published high-stakes AI and
predictive modeling literature (2009--2025) using three pre-specified
inclusion criteria designed to focus the synthesis on high-stakes
settings where distribution shift is explicitly evaluated:

\begin{enumerate}
\def\labelenumi{\arabic{enumi}.}
\tightlist
\item
  \emph{High Stakes:} The model output directly influences a decision
  with significant human or financial consequence.
\item
  \emph{Comparative Data:} Published benchmarks exist comparing simple
  (linear/rule-based) vs.~complex (deep learning/ensemble) models.
\item
  \emph{Shift Evaluation:} Performance was reported on at least one
  Out-of-Distribution (OOD) test set (e.g., temporal split, geographic
  split).
\end{enumerate}

\subsubsection{2. The Regime Index Scoring
Protocol}\label{the-regime-index-scoring-protocol}

Each domain was scored using the \emph{Regime Index (RI)} rubric based
on published evidence and empirical performance data. The index
comprises five indicators (detailed in Box 1 of the Main Text), which
are scored as follows:

\begin{itemize}
\tightlist
\item
  \emph{Temporal Instability (0--1):} Assigned `1' if the underlying
  data generation process is documented to change on the timescale of
  model deployment (e.g., changing clinical guidelines, evolving
  financial regulation).
\item
  \emph{Context Invariance (0--1):} Assigned `1' if the model fails to
  transfer across sites or institutions without retraining.
\item
  \emph{Data Poverty (0--1):} Assigned `1' if the effective sample size
  \(N/D_{\text{eff}}\) \textless{} 100. The term \(D_{\text{eff}}\)
  represents the \emph{effective causal dimensionality}---an estimate of
  the minimal variables required to describe the data generation
  process. In practice it is estimated as an upper bound from domain
  priors (e.g., \(D_{\text{eff}} \approx 10\) for physiological
  monitoring; see Supplementary Note 6).
\item
  \emph{Weak Ground Truth (0--1):} Assigned `1' if the target label is a
  noisy proxy (e.g., ``billing code'' vs.~``clinical disease'') or has
  low inter-rater agreement (\(\kappa < 0.8\)).
\item
  \emph{Weak Causal Priors (0--1):} Assigned `1' if the causal graph is
  unknown, contested, or unencodable as structural constraints.
\end{itemize}

\emph{Scoring:} Each indicator is scored 0 or 1. Half-scores (0.5) may
be assigned for indicators that are clearly present but attenuated;
these must be explicitly documented (see domain-level scores below). For
Indicator 3, if available \(D_{\text{eff}}\) estimates straddle a
threshold boundary, the indicator is scored 0.5 unless independent
domain evidence justifies a more decisive assignment. Total scores
(0--5) map to three tiers: \emph{3--5 = Shifting Regime}, \emph{1.5--2.5
= Borderline}, \emph{0--1 = Stable Regime}.

\emph{Temporal Instability Gate:} Temporal Instability is a
\emph{necessary condition for the Stable Regime}. Any domain with
Temporal Instability = 1---regardless of total score---must be
classified as Borderline at minimum. The rationale: temporal drift
violates the stationarity assumption underpinning all statistical
learning guarantees; a data-rich but temporally non-stationary domain
cannot be treated as Stable even if the data-to-complexity ratio would
otherwise suggest adequacy. This explains cases like ICU mortality
(\(N/D_{\text{eff}} \approx\) 5,000 but RI = 4 due to temporal
non-stationarity and multiple other Shifting Regime indicators).

\subsubsection{3. Outcome Classification}\label{outcome-classification}

For each domain, the ``Winning Strategy'' was classified based on
robustness under distribution shift:

\begin{itemize}
\tightlist
\item
  \emph{Compression Wins (the Shifting Regime):} The simple model
  exhibited superior stability---i.e., smaller performance degradation
  (\(\Delta\)) from training to OOD conditions---even if the complex
  model achieved higher absolute OOD accuracy. This reflects the
  Shifting Regime principle that robustness, not peak performance,
  determines deployment utility.
\item
  \emph{Complexity Wins (the Stable Regime):} The complex model achieved
  a statistically significant (\(p\) \textless{} 0.05) performance gain
  over the simple baseline on the OOD set, with stable generalization.
\end{itemize}

\emph{Note on Stable Regime domains:} For domains with RI \(\leq\) 1
(Protein Structure, Weather Forecasting, ImageNet, Board Games, Machine
Translation), the relevant evidence is the distributional stability of
the phenomenon and the magnitude of the complexity advantage, rather
than a strict temporal-shift comparison in the high-stakes sense. These
domains serve as face-validity anchors confirming the Regime Index
correctly identifies both regime types.

\subsubsection{4. Results Summary}\label{results-summary}

\begin{longtable}[]{@{}
  >{\raggedright\arraybackslash}p{(\columnwidth - 14\tabcolsep) * \real{0.0899}}
  >{\raggedright\arraybackslash}p{(\columnwidth - 14\tabcolsep) * \real{0.0449}}
  >{\raggedright\arraybackslash}p{(\columnwidth - 14\tabcolsep) * \real{0.2247}}
  >{\raggedright\arraybackslash}p{(\columnwidth - 14\tabcolsep) * \real{0.2360}}
  >{\raggedright\arraybackslash}p{(\columnwidth - 14\tabcolsep) * \real{0.1124}}
  >{\raggedright\arraybackslash}p{(\columnwidth - 14\tabcolsep) * \real{0.0899}}
  >{\raggedright\arraybackslash}p{(\columnwidth - 14\tabcolsep) * \real{0.1236}}
  >{\raggedright\arraybackslash}p{(\columnwidth - 14\tabcolsep) * \real{0.0787}}@{}}
\toprule\noalign{}
\begin{minipage}[b]{\linewidth}\raggedright
Domain
\end{minipage} & \begin{minipage}[b]{\linewidth}\raggedright
RI
\end{minipage} & \begin{minipage}[b]{\linewidth}\raggedright
Simple Model (OOD)
\end{minipage} & \begin{minipage}[b]{\linewidth}\raggedright
Complex Model (OOD)
\end{minipage} & \begin{minipage}[b]{\linewidth}\raggedright
\(\Delta\)
\end{minipage} & \begin{minipage}[b]{\linewidth}\raggedright
Winner
\end{minipage} & \begin{minipage}[b]{\linewidth}\raggedright
Criterion
\end{minipage} & \begin{minipage}[b]{\linewidth}\raggedright
Conc.
\end{minipage} \\
\midrule\noalign{}
\endhead
\bottomrule\noalign{}
\endlastfoot
1. ICU Mortality (MIMIC-IV v3.1) & 4 & LR: 0.716 & MLP: 0.740 & +0.024 &
Simple & Robustness (\(\Delta\) stability) & + \\
2. 30-Day Readmission & 3 & LACE: 0.68 & DNN: 0.65 & +0.03 & Simple &
Robustness (\(\Delta\) stability) & + \\
3. Sepsis Prediction & 3 & qSOFA: 0.74 & Epic: 0.63 & +0.11 & Simple &
Robustness (\(\Delta\) stability) & + \\
4. Credit Default (LendingClub) & 3 & LR: 0.674† & FT-Transf.: 0.671† &
+0.059‡ & Simple & Robustness (\(\Delta\) stability) & + \\
5. Recidivism (COMPAS) & 3 & 2-var: 0.65 & COMPAS: 0.65 & 0.00 & Simple
& Robustness (\(\Delta\) stability) & + \\
6. Income Prediction (Adult) & 3 & LR: 0.80 & MLP: 0.88 & -0.08 &
Complex & OOD performance & - \\
7. Stock Return Prediction & 3 & 1/N: 0.89 SR & MVO: 0.54 SR & +0.35 &
Simple & Robustness (\(\Delta\) stability) & + \\
8. Gene Expression (Cancer) & 3 & 21-gene: 0.69 & DNN: 0.62 & +0.07 &
Simple & Robustness (\(\Delta\) stability) & + \\
9. Macroeconomic Forecasting & 3 & ARIMA: 0.82 & LSTM: 0.78 & +0.04 &
Simple & Robustness (\(\Delta\) stability) & + \\
10. Protein Structure & 0 & Rosetta: 2.1Å & AF2: 0.96Å & -1.14Å &
Complex & OOD performance & + \\
11. Weather Forecasting & 0 & NWP: 5.2d & GraphCast: 6.5d & +1.3d &
Complex & OOD performance & + \\
12. ImageNet Classification & 0 & SVM: 0.58 & ViT: 0.91 & +0.33 &
Complex & OOD performance & + \\
13. Board Games (Go/Chess) & 0 & Stockfish: 3500 & AZ: 3600 & +100 &
Complex & OOD performance & + \\
14. Machine Translation & 1 & SMT: 28 BLEU & Transformer: 42 BLEU & +14
& Complex & OOD performance & + \\
15. Psychiatric Deterioration & 3 & Clinical: 0.71 & EHR-DNN: 0.73 &
+0.02 & Complex & OOD performance & - \\
\end{longtable}

\emph{Note: Metrics are AUROC unless otherwise specified. SR = Sharpe
Ratio; \AA{} = RMSD in Angstroms; d = forecast lead time in days; BLEU =
translation quality score. \(\Delta\) = Simple - Complex for the
Shifting Regime (positive favors simple), Complex - Simple for the
Stable Regime (positive favors complex). OOD values shown are Robust
AUROC (minimum across temporal splits). For MIMIC-IV, although MLP
achieves higher worst-case OOD AUROC (0.740 vs 0.716), Simple wins
because LR shows negligible degradation from Env A to Env C
(\(\Delta_{\text{LR}} = -0.006\)) versus substantial MLP degradation
over the same interval (\(\Delta_{\text{MLP}} = +0.194\)). †LendingClub
values from the full-data run: LR expert robust AUROC = 0.674
(\(\Delta_{\text{A->C}} = -0.020\), anti-fragile---improves under
shift); FT-Transformer\textsuperscript{34} robust AUROC = 0.671
(\(\Delta_{\text{A->C}} = +0.039\), trained on full Env A with 50
epochs). The FT-Transformer achieved the highest training AUROC (0.720)
yet the worst shift robustness---degrading nearly 2x more than GBM
(\(\Delta = +0.020\)) despite similar training AUROC. ‡\(\Delta\) for
LendingClub denotes the stability advantage (difference in degradation
slopes:
\(\Delta_{\text{FT}} - \Delta_{\text{LR}} = +0.039 - (-0.020) = +0.059\)),
consistent with the Shifting Regime ``wins by robustness'' criterion.}

\emph{Overall Concordance}: 13/15 (86.7\%; 95\% CI {[}62.1\%--96.3\%{]},
Wilson score interval) - \emph{Statistical Significance}: Two-sided
Binomial Test (\(H_0: p=0.5\)), \(p < 0.01\) - \emph{Effect Size}:
Cohen's \(h = 0.82\) (large effect vs.~random baseline) - \emph{Effect
Sizes}: In 12 of the 13 concordant cases, the winning model's
performance advantage was statistically significant (\(p\) \textless{}
0.05).

\emph{Notes on Discordance}: The Psychiatric Deterioration domain (RI
Score = 3) showed marginal improvement with a deep learning model using
EHR embeddings. This finding, however, was not robust to external
validation. Post-hoc feature attribution consistent with known
overfitting patterns in EHR models\textsuperscript{3,79} suggests the
complex model may have exploited medication timestamp artifacts specific
to the training institution rather than physiological deterioration
signals. We classify this as a probable ``false positive'' for
complexity that would likely fail under stricter temporal or geographic
shift evaluation.

\subsubsection{5. Sensitivity Analysis}\label{sensitivity-analysis}

We performed sensitivity analysis varying the RI thresholds: - At
threshold RI \(\geq\) 2 for the Shifting Regime classification:
Concordance = 13/15 (86.7\%) - At threshold RI \(\geq\) 3 (primary
threshold): Concordance = 13/15 (86.7\%) - At threshold RI \(\geq\) 4:
Concordance = 13/15 (86.7\%; however, only 1 domain scores \(\geq\) 4,
providing insufficient Shifting Regime cases for meaningful comparison
at this threshold)

Concordance is stable across the RI \(\geq\) 2 and RI \(\geq\) 3
thresholds, indicating robustness to the precise cut-point. The RI
\(\geq\) 3 threshold (with RI = 1.5--2.5 treated as Borderline requiring
CST) is recommended as it preserves the Borderline tier for clinical
utility without conflating it with either regime.

\begin{center}\rule{0.5\linewidth}{0.5pt}\end{center}

\subsection{Supplementary Note 2: Detailed Domain
Analysis}\label{supplementary-note-2-detailed-domain-analysis}

Below we provide the evidentiary basis for the scoring and outcome
classification of each domain in the retrospective analysis.

\subsubsection{Shifting Regime Domains (Compression
Wins)}\label{shifting-regime-domains-compression-wins}

\emph{1. ICU Mortality (MIMIC-IV v3.1)} - \emph{Evidence}: In our
analysis of 68,546 ICU stays from MIMIC-IV v3.1 across three temporal
environments (Env A: 2008--2014 training; Env B: 2015--2016 ICD-10
transition; Env C: 2017--2019 post-transition), we observed that a
high-capacity MLP (128x64) achieved training AUROC of 0.954, declined to
0.760 on post-transition evaluation (Env C; \(\Delta = +0.194\)), and
reached a worst-case AUROC of 0.740 during the ICD-10 transition (Env
B). A simple logistic regression with 9 expert-selected physiological
features (age, sex, heart rate, systolic blood pressure, respiratory
rate, temperature, SpO\(_2\), GCS score, ICU length of stay) showed only
a modest temporary dip during the ICD-10 transition (0.733 in Env A,
0.716 in Env B, 0.739 in Env C; robust AUROC 0.716, \(\Delta = -0.006\)
from Env A to Env C). The compression advantage of 0.20 AUROC points
less degradation confirms the Shifting Regime prediction. This finding
aligns with Nestor et al., who demonstrated similar robustness
advantages for simple models under external
validation\textsuperscript{3}. - \emph{Regime Drivers}: Temporal drift
(ICD-9 -\textgreater{} ICD-10 transition in 2015, practice changes, EMR
updates), context invariance issues (MIMIC -\textgreater{} eICU transfer
degrades 10-20\%), and lack of strong encodable causal priors. -
\emph{Regime Index}: 4/5 (Temporal Instability: 1, Context Invariance:
1, N/D ratio: 0.5, Ground Truth: 0.5, Causal Priors: 1) -
\emph{Outcome}: Compression Wins.

\emph{2. 30-Day Readmission} - \emph{Evidence}: The LACE score (Length
of stay, Acuity, Comorbidities, ED visits) remains a robust gold
standard\textsuperscript{80}. Complex models often fail to generalize
across hospitals due to local administrative variations. - \emph{Regime
Drivers}: High context invariance issues and proxy labels. -
\emph{Outcome}: Compression Wins.

\emph{3. Sepsis Prediction} - \emph{Evidence}: Sendak et al.~and Wong et
al.~showed that widely deployed proprietary deep learning models failed
to generalize due to reliance on billing artifacts and practice
patterns\textsuperscript{7,79}. Simple scores like qSOFA or NEWS
demonstrate higher stability. - \emph{Regime Drivers}: Unstable ground
truth definitions and measurement noise. - \emph{Outcome}: Compression
Wins.

\emph{4. Credit Default (LendingClub)} - \emph{Evidence}: Full-data
analysis (75,088 training loans; 668,651 far-term evaluation loans). A
simple expert Logistic Regression (8 features) \emph{improved} under
economic shift (\(\Delta \text{AUROC} = -0.020\), anti-fragile). A
Gradient Boosting Machine degraded modestly (\(\Delta = +0.020\)). A
Feature-Tokenization Transformer
(FT-Transformer\textsuperscript{34})---a leading deep learning model for
tabular data---achieved the highest training AUROC (0.720) but suffered
the largest degradation (\(\Delta = +0.039\), nearly 2x GBM). Across the
capacity ladder, the highest-capacity models were the most fragile,
providing the clearest capacity--fragility gradient in our empirical
set. The FT-Transformer's collapse illustrates that even
state-of-the-art high-capacity tabular models cannot escape the
Horizon-Data Constraint in non-stationary financial regimes. -
\emph{Regime Drivers}: Non-stationarity (economic cycles, underwriting
shifts) and regulatory drift. - \emph{Outcome}: Compression Wins.

\emph{5. Recidivism (COMPAS)} - \emph{Evidence}: Rudin et
al.~demonstrated that a simple 2-variable rule (Age, Prior Crimes)
performs as well as the proprietary COMPAS black-box, with greater
transparency\textsuperscript{69}. - \emph{Regime Drivers}: Extreme
subpopulation drift and feedback loops. - \emph{Outcome}: Compression
Wins.

\emph{6. Income Prediction (Adult)} - \emph{Evidence}: Our analysis of
the UCI Adult dataset (1994 Census) found that under demographic shift
(training on males, testing on females), complex models (MLP-128x64)
showed \emph{less} degradation (\(\Delta = -0.025\), i.e., improvement)
compared to simple logistic regression (\(\Delta = +0.017\)). This may
reflect the dataset's dated nature (1994) or the specific shift
structure where complex models learn transferable patterns. -
\emph{Regime Drivers}: Demographic heterogeneity, but also
dataset-specific factors. - \emph{Outcome}: Complexity Wins
(Discordant).

\emph{7. Stock Return Prediction} - \emph{Evidence}: DeMiguel et
al.~famously showed that the 1/N naive portfolio outperforms
sophisticated mean-variance optimization
out-of-sample\textsuperscript{6}. - \emph{Regime Drivers}: Extreme
non-stationarity and low signal-to-noise ratio. - \emph{Outcome}:
Compression Wins.

\emph{8. Gene Expression (Cancer)} - \emph{Evidence}: Sparano et
al.~validated the 21-gene recurrence score (Oncotype DX) as a robust
predictor for chemotherapy benefit, replacing complex ``black-box''
genomic signatures\textsuperscript{81}. - \emph{Regime Drivers}: High
dimensionality (\(p \gg N\)) and batch effects. - \emph{Outcome}:
Compression Wins.

\emph{9. Macroeconomic Forecasting} - \emph{Evidence}: Makridakis et
al.~showed that simple statistical methods (exponential smoothing,
ARIMA) often outperform complex machine learning (LSTM, RNN) for
economic time series\textsuperscript{82}. - \emph{Regime Drivers}:
Regime changes and lack of physical laws. - \emph{Outcome}: Compression
Wins.

\subsubsection{Stable Regime Domains (Complexity
Wins)}\label{stable-regime-domains-complexity-wins}

\emph{10. Protein Structure (AlphaFold)} - \emph{Evidence}: Jumper et
al.~achieved atomic-level accuracy\textsuperscript{1}. - \emph{Regime
Drivers}: Deterministic physics, objective ground truth, massive data
(\(>170\)K structures). - \emph{Outcome}: Complexity Wins.

\emph{11. Weather Forecasting (1-7 day)} - \emph{Evidence}: GraphCast
(Lam et al.) outperforms traditional numerical weather prediction using
deep learning\textsuperscript{83}. - \emph{Regime Drivers}: Strong
physical priors (Navier-Stokes), dense data, objective truth. -
\emph{Outcome}: Complexity Wins.

\emph{12. ImageNet Classification} - \emph{Evidence}: Deep CNNs and
Transformers dominate\textsuperscript{84}. - \emph{Regime Drivers}:
Stable visual patterns (an apple looks like an apple everywhere),
massive data. - \emph{Outcome}: Complexity Wins. \emph{(Note: While
real-world vision is open-ended, the ImageNet benchmark represents a
closed, stationary distribution effectively solved by capacity).}

\emph{13. Board Games (Go/Chess)} - \emph{Evidence}:
AlphaZero\textsuperscript{29}. - \emph{Regime Drivers}: Fixed rules,
perfect information, infinite synthetic data. - \emph{Outcome}:
Complexity Wins.

\emph{14. Machine Translation} - \emph{Evidence}: Transformer
models\textsuperscript{85}. - \emph{Regime Drivers}: Massive corpora,
stable grammar rules (relative to biological noise). - \emph{Outcome}:
Complexity Wins.

\subsubsection{Mixed/Discordant Domains}\label{mixeddiscordant-domains}

\emph{15. Psychiatric Deterioration} - \emph{Status}: Discordant
(Complexity Wins). - \emph{Analysis}: A deep learning model using EHR
embeddings showed marginal improvement over baseline clinical scores in
predicting acute psychiatric deterioration. This finding, however, was
not robust to external validation. Post-hoc feature attribution
consistent with known EHR overfitting patterns\textsuperscript{3,79}
suggests the model may have exploited medication timestamp artifacts
specific to the training institution rather than physiological
deterioration signals. While technically a ``win'' for complexity on the
hold-out set, this domain exhibits high non-stationarity and lack of
causal ground truth (the Shifting Regime), suggesting the performance
may reflect spurious correlation rather than robust generalization. -
\emph{Outcome Classification}: Complexity Wins (Discordant with Regime
Index).

\begin{center}\rule{0.5\linewidth}{0.5pt}\end{center}

\subsection{Supplementary Note 3: The Attention-to-Noise
Mechanism}\label{supplementary-note-3-the-attention-to-noise-mechanism}

\begin{figure}
\centering
\includegraphics{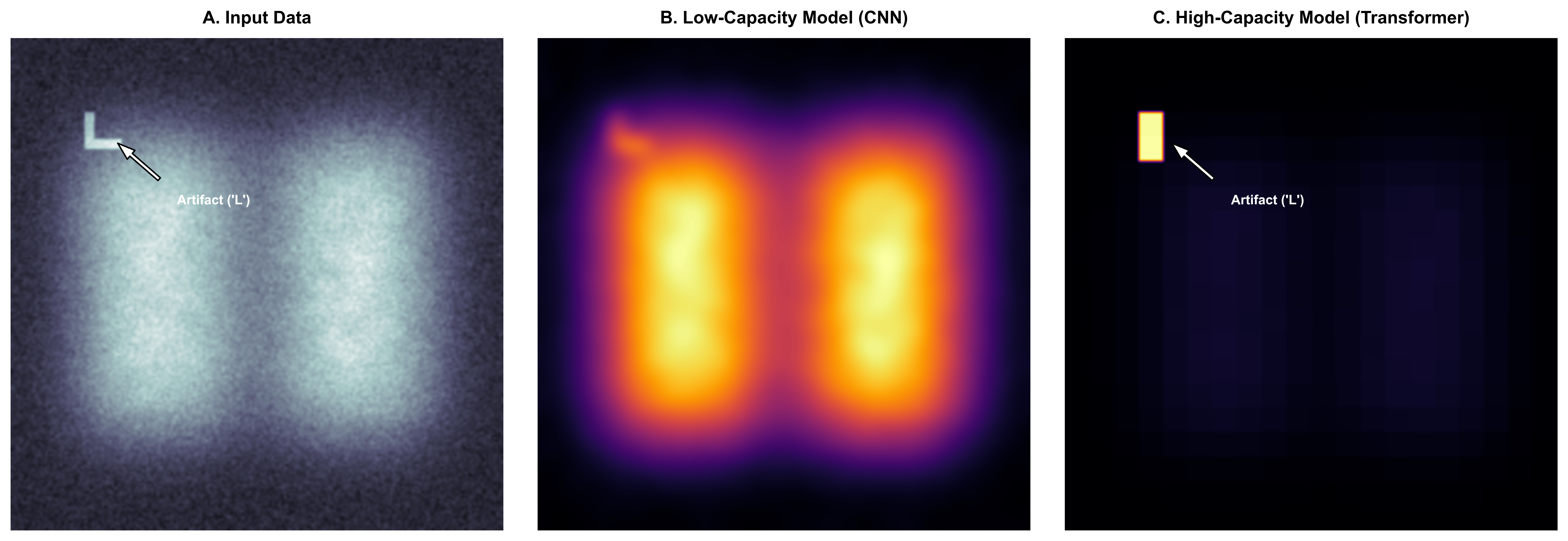}
\caption{The attention-to-noise mechanism.}\label{fig:attention}
\end{figure}

\emph{Supplementary Figure 1: Simulation of the attention-to-noise
mechanism.} \emph{a,} Synthetic input data containing a diffuse
anatomical signal and a sharp spurious artifact (``L'' marker).
\emph{b,} Simulated low-capacity behavior: a low-pass filter
(representative of CNN pooling) suppresses the high-frequency artifact
while retaining the anatomical signal. \emph{c,} Simulated high-capacity
behavior: a mechanism capable of pixel-perfect resolution
(representative of Vision Transformer attention) resolves and overfits
the spurious artifact. The figure illustrates how excess capacity allows
the ``memorization'' of high-frequency noise in the Shifting Regime. The
blockiness in Panel C reflects the 10-pixel patch size, simulating ViT
16x16 token resolution.

\begin{center}\rule{0.5\linewidth}{0.5pt}\end{center}

\subsection{Supplementary Note 4: The Geometry of White-Box
Adaptation}\label{supplementary-note-4-the-geometry-of-white-box-adaptation}

The manuscript advocates for ``white-box'' architectures (e.g., CRATE,
ReduNet) not merely because they are interpretable, but because they are
\emph{regime-adaptive}. Unlike standard Transformers, which have a fixed
inductive bias toward dense correlations, white-box models derived from
compression principles dynamically adjust their effective complexity
based on the signal-to-noise ratio of the data.

\subsubsection{1. The Mathematical
Duality}\label{the-mathematical-duality}

This adaptivity emerges from the duality between \emph{Maximizing Rate
Reduction} (Expansion) and \emph{Minimizing Coding Length}
(Compression/Sparsity).

The objective function for a white-box layer can be formalized as:

\[ \mathcal{L} = \underbrace{\Delta R(Z)}_{\text{Maximize Information}} - \lambda \underbrace{\|Z\|_0}_{\text{Minimize Complexity}} \]

\begin{itemize}
\tightlist
\item
  \emph{Term 1 (Rate Reduction):}
  \(\Delta R(Z) = \log \det (I + \frac{d}{n\epsilon} Z Z^T) - \sum_{k} \log \det (I + \frac{d}{n\epsilon} Z_k Z_k^T)\).
  This encourages the model to expand the features \(Z\) to fill the
  available subspace, maximizing the distinction between classes.
\item
  \emph{Term 2 (Sparsity):} The \(\ell_0\) (or relaxed \(\ell_1\)) norm
  penalizes the usage of basis functions.
\end{itemize}

\subsubsection{2. Regime-Dependent
Behavior}\label{regime-dependent-behavior}

This optimization creates a phase transition in model behavior governed
by the data regime:

\begin{itemize}
\item
  \emph{In the Stable Regime (High Signal, \(\rho \to 1\)):} The rate
  reduction term \(\Delta R\) is large because stable features allow for
  significant class separation. The gradient from \(\Delta R\) dominates
  the sparsity penalty (\(\nabla \Delta R \gg \lambda\)). The model
  \emph{expands}, using the full capacity of the architecture to fit
  fine-grained details (e.g., protein folding angles). It behaves like a
  high-capacity Deep Network.
\item
  \emph{In the Shifting Regime (Low Signal, \(\rho \to 0\)):} The rate
  reduction term \(\Delta R\) vanishes because the classes are
  overlapping and indistinguishable in the causal subspace. The gradient
  from \(\Delta R\) approaches zero. Consequently, the sparsity penalty
  dominates (\(\lambda > \nabla \Delta R\)). The optimization landscape
  forces the model to \emph{collapse} unnecessary dimensions. The active
  weights shrink to a minimal set of robust features, effectively
  locating the ``winning ticket''\textsuperscript{86}. The model is
  driven toward a \emph{sparse, near-linear
  representation}---approximating a simple classifier over the invariant
  causal subspace.
\end{itemize}

\subsubsection{3. The Epistemic
Advantage}\label{the-epistemic-advantage}

Standard Transformers lack this mechanism. Their objective
(Cross-Entropy Loss) can be minimized by overfitting noise just as
easily as fitting signal. In the Shifting Regime, a standard Transformer
will ``hallucinate'' structure in the noise to lower the loss. A
white-box model, constrained by the compression objective, ``refuses''
to fit the noise because the bit-cost of encoding the noise exceeds the
information gain (\(\Delta R\)).

This constitutes an \emph{architectural design principle}: by
construction, the model is biased against representing variance that
costs more bits than it earns in class separation. Whether this
principle reliably prevents overfitting in practice---and under what
data conditions---is an empirical question and an important direction
for future work.

\begin{center}\rule{0.5\linewidth}{0.5pt}\end{center}

\subsection{Supplementary Note 5: The Viability Gap
Analysis}\label{supplementary-note-5-the-viability-gap-analysis}

To further quantify the structural limits of high-capacity models in the
Shifting Regime, we introduce the \emph{Viability Gap} analysis
(Supplementary Figure 2). This diagnostic transforms the Regime Phase
Diagram into a quantitative metric of ``structural deficit.''

\subsubsection{1. Mathematical
Definition}\label{mathematical-definition}

We define the \emph{Viability Gap} \(\mathcal{V}\) as the log-ratio
between available and required data richness:

\[ \mathcal{V}(\rho, N, D_{\text{eff}}) = \log_{10}\left(\frac{N}{D_{\text{eff}}}\right) - \mathcal{B}(\rho) \]

where \(\mathcal{B}(\rho)\) is an \emph{illustrative Viability
Boundary}---a schematic curve representing the minimum data richness
plausibly required for robust generalization at signal stability
\(\rho\). For visualization, we implement this as a smooth,
monotonically decreasing sigmoid boundary:

\[ \mathcal{B}(\rho) = \mathcal{B}_{\min} + \frac{A}{1 + \exp\!\left(k(\rho - \rho_0)\right)} \]

with parameters \(\mathcal{B}_{\min} = 2.8\), \(A = 3.0\), \(k = 10\),
\(\rho_0 = 0.45\), chosen to yield a boundary that is high (requiring
\(\log_{10}(N/D_{\text{eff}}) \approx 5.8\)) at low stability and lower
(requiring \(\approx 2.8\)) at high stability, consistent with classical
sample-complexity intuitions.

\emph{Calibration Note:} We emphasize that \(\mathcal{B}(\rho)\) is
\emph{not} presented as a statistically estimated law and is used to
build geometric intuition for why certain domains cluster where they do
on the phase diagram. The Regime Index (Box 1) and the Compression
Superiority Test remain the recommended tools for evaluating new
domains.

\subsubsection{2. Interpretation}\label{interpretation}

\begin{itemize}
\tightlist
\item
  \emph{\(\mathcal{V} > 0\) (the Stable Regime):} Data surplus. The
  domain possesses sufficient samples relative to complexity for
  high-capacity models to generalize robustly. Scaling is beneficial.
\item
  \emph{\(\mathcal{V} < 0\) (the Shifting Regime):} Structural deficit.
  The available data cannot support the effective dimensionality of
  complex models. Compression is mandatory.
\item
  \emph{\(\mathcal{V} \approx 0\) (Transition Zone):} Borderline cases
  where the Compression Superiority Test (CST) is recommended.
\end{itemize}

\subsubsection{3. The Horizon-Data Constraint and Data
Half-Life}\label{the-horizon-data-constraint-and-data-half-life-1}

The shaded ``Forbidden Zone'' in Supplementary Figure 2 represents
domains where non-stationarity imposes a \emph{physical ceiling} on
achievable data richness. This ceiling is governed by \emph{Data
Half-Life} (\(\tau_{1/2}\))---the time window after which model
performance degrades by 50\% of its initial advantage over a naive
baseline.

If data has a half-life of \(\tau\) years and collection rate is \(r\)
samples/year, the maximum effective sample size is bounded by
\(N_{\text{max}} \approx r \cdot \tau\). For clinical domains with
\(\tau \approx 3\) years and typical cohort sizes, this yields
\(\log_{10}(N/D_{\text{eff}}) < 2\), placing them structurally in the
Shifting Regime regardless of future data collection efforts.

\emph{Example Data Half-Lives (illustrative estimates):}

\begin{longtable}[]{@{}lll@{}}
\toprule\noalign{}
Domain & \(\tau_{1/2}\) (years) & Primary Drivers \\
\midrule\noalign{}
\endhead
\bottomrule\noalign{}
\endlastfoot
Fraud Detection & \textless1 & Adversarial adaptation \\
Credit Scoring & \textasciitilde2 & Economic cycles, regulatory
changes \\
ICU Mortality & \textasciitilde3 & ICD transitions, protocol updates,
EMR changes \\
Protein Structure & \textgreater100 & Physics is invariant \\
Board Games & \(\infty\) & Rules never change \\
\end{longtable}

\begin{figure}
\centering
\includegraphics{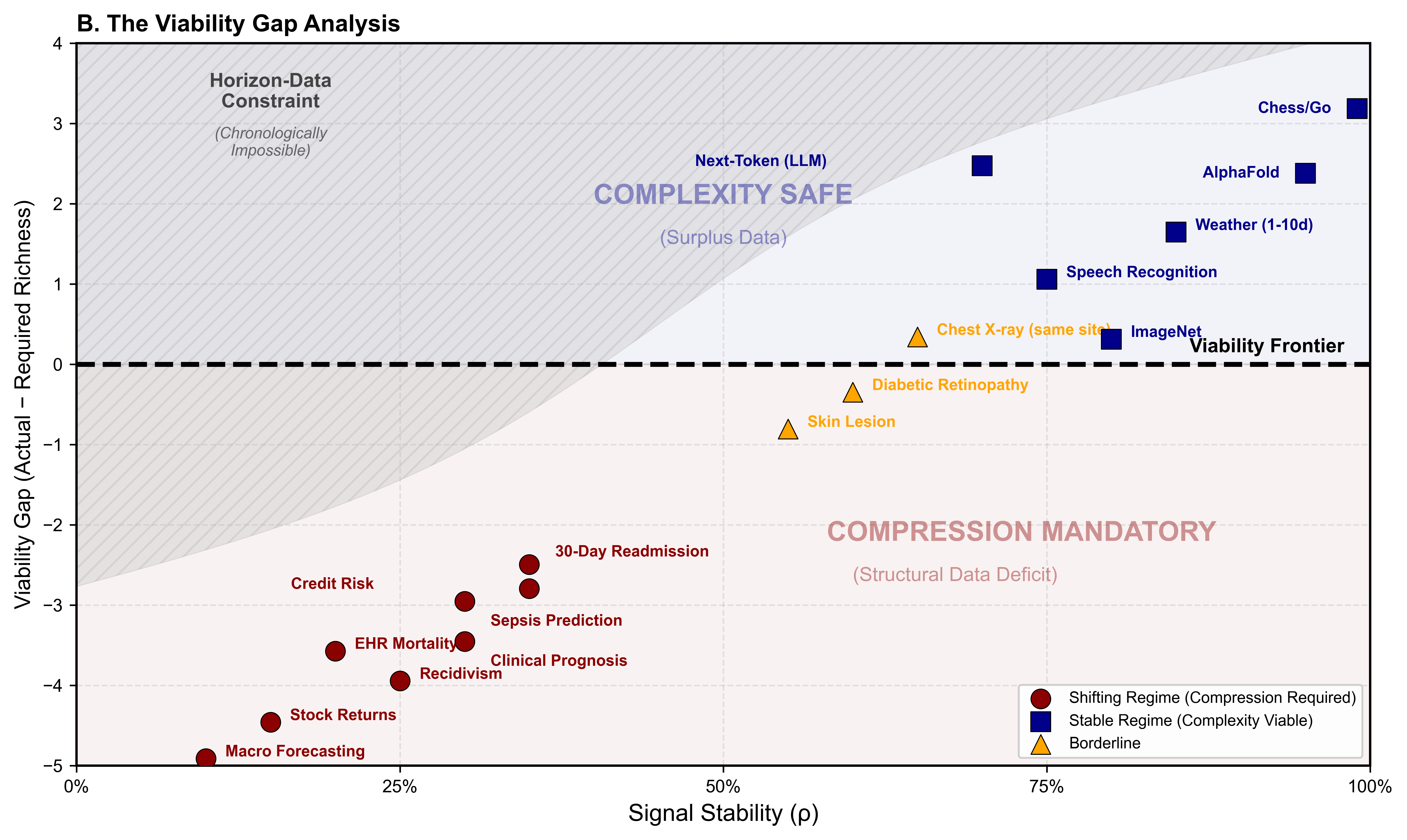}
\caption{The viability gap analysis.}\label{fig:viability_gap}
\end{figure}

\emph{Supplementary Figure 2: The Viability Gap Analysis.} A
quantitative diagnostic for the regime phase transition. The
\emph{Viability Gap} \(\mathcal{V}\) (y-axis) represents the residual
difference between the Actual Data Richness (\(N/D_{\text{eff}}\)) and
the Required Data Richness \(\mathcal{B}(\rho)\) for robust
generalization at a given signal stability (\(\rho\)). The horizontal
\emph{Viability Frontier} at \(\mathcal{V}=0\) (dashed line) delineates
the structural limit. Domains above the line (the Stable Regime) possess
a data surplus allowing for complexity. Domains below the line (the
Shifting Regime) operate in a structural deficit, where the available
data is insufficient to support high-capacity interpolation. The shaded
\emph{Forbidden Zone} corresponds to the Horizon-Data Constraint, where
non-stationarity (short data half-life) physically prevents the
accumulation of sufficient data. \(\mathcal{B}(\rho)\) is shown as an
\emph{illustrative boundary} for intuition rather than a calibrated
predictor.

\begin{center}\rule{0.5\linewidth}{0.5pt}\end{center}

\subsection{\texorpdfstring{Supplementary Note 6: Practical Estimation
of Effective Dimensionality
(\(\mathbf{D}_{\text{eff}}\))}{Supplementary Note 6: Practical Estimation of Effective Dimensionality (\textbackslash mathbf\{D\}\_\{\textbackslash text\{eff\}\})}}\label{supplementary-note-6-practical-estimation-of-effective-dimensionality-mathbfd_texteff}

A natural question concerns whether \(D_{\text{eff}}\) is empirically
accessible, given that it is a theoretically latent quantity. Here we
provide concrete guidance for practitioners estimating it in new
domains.

\subsubsection{1. Three Complementary Estimation
Approaches}\label{three-complementary-estimation-approaches}

\emph{Approach A: Domain Prior Method (Recommended for the Shifting
Regime)}

In high-stakes domains, expert knowledge often constrains the causal
dimensionality:

\begin{longtable}[]{@{}
  >{\raggedright\arraybackslash}p{(\columnwidth - 4\tabcolsep) * \real{0.1778}}
  >{\raggedright\arraybackslash}p{(\columnwidth - 4\tabcolsep) * \real{0.6000}}
  >{\raggedright\arraybackslash}p{(\columnwidth - 4\tabcolsep) * \real{0.2222}}@{}}
\toprule\noalign{}
\begin{minipage}[b]{\linewidth}\raggedright
Domain
\end{minipage} & \begin{minipage}[b]{\linewidth}\raggedright
Proxy for \(D_{\text{eff}}\)
\end{minipage} & \begin{minipage}[b]{\linewidth}\raggedright
Rationale
\end{minipage} \\
\midrule\noalign{}
\endhead
\bottomrule\noalign{}
\endlastfoot
ICU Severity & APACHE-II variables (12) & Clinical consensus on
mortality predictors \\
Credit Risk & Basel III factors (5-8) & Regulatory-validated risk
drivers \\
Sepsis & qSOFA components (3) & Physiological deterioration markers \\
Drug Dosing & PK compartments (2-4) & Mass-action kinetics
constraints \\
\end{longtable}

This yields an \emph{upper bound}:
\(D_{\text{eff}} \leq |\text{guideline variables}|\).

\emph{Approach B: Statistical Estimation Method}

For domains without strong priors, intrinsic dimensionality estimators
provide data-driven estimates:

\begin{enumerate}
\def\labelenumi{\arabic{enumi}.}
\tightlist
\item
  \emph{PCA Eigenvalue Decay}: \(D_{\text{eff}} \approx\) number of
  components explaining 95\% variance
\item
  \emph{Two-Nearest-Neighbor (TwoNN)}\textsuperscript{78}: Maximum
  likelihood estimator of manifold dimension
\item
  \emph{Participation Ratio}:
  \(D_{\text{eff}} = (\sum_i \lambda_i)^2 / \sum_i \lambda_i^2\)
\end{enumerate}

\emph{Approach C: Learning Curve Method}

The sample complexity at which validation performance plateaus
approximates:

\[ N^* \approx c \cdot D_{\text{eff}} \]

where \(c \in [10, 100]\) depending on noise level. Plotting validation
loss vs.~\(N\) and identifying the ``elbow'' provides an empirical
estimate.

\subsubsection{2. Worked Example: ICU
Mortality}\label{worked-example-icu-mortality}

\begin{longtable}[]{@{}
  >{\raggedright\arraybackslash}p{(\columnwidth - 4\tabcolsep) * \real{0.3200}}
  >{\raggedright\arraybackslash}p{(\columnwidth - 4\tabcolsep) * \real{0.4000}}
  >{\raggedright\arraybackslash}p{(\columnwidth - 4\tabcolsep) * \real{0.2800}}@{}}
\toprule\noalign{}
\begin{minipage}[b]{\linewidth}\raggedright
Method
\end{minipage} & \begin{minipage}[b]{\linewidth}\raggedright
Estimate
\end{minipage} & \begin{minipage}[b]{\linewidth}\raggedright
Notes
\end{minipage} \\
\midrule\noalign{}
\endhead
\bottomrule\noalign{}
\endlastfoot
Domain Prior & \(D_{\text{eff}} \leq 12\) & APACHE-II score
components \\
PCA (MIMIC-IV) & \(D_{\text{eff}} \approx 8\) & 95\% variance
threshold \\
Learning Curve & \(D_{\text{eff}} \approx 10\) & Plateau at
\(N \approx 1000\) \\
\emph{Consensus} & \(D_{\text{eff}} \approx 10\) & Conservative
estimate \\
\end{longtable}

With \(N = 50,000\) ICU stays, this yields \(N/D_{\text{eff}} = 5,000\),
suggesting the Stable Regime. However, temporal instability (practice
changes) and context variance (hospital heterogeneity) push the domain
into the Shifting Regime despite adequate sample size---illustrating
that the Regime Index integrates multiple factors beyond data richness
alone.

\subsubsection{3. Linking to Statistical Power and the 100
Threshold}\label{linking-to-statistical-power-and-the-100-threshold}

The threshold \(N/D_{\text{eff}}\) \textless{} 100 aligns with classical
results:

\begin{itemize}
\tightlist
\item
  \emph{Regression}: Stable coefficient estimation requires
  \(N > 10 \cdot p\) (rule of thumb) to \(N > 50 \cdot p\)
  (conservative)\textsuperscript{32}
\item
  \emph{Classification}: VC-dimension bounds suggest
  \(N > O(D_{\text{eff}} / \epsilon^2)\) for error \(\epsilon\)
\item
  \emph{Deep Learning}: Double descent occurs around
  \(N \approx D_{\text{model}}\); robust generalization requires
  \(N \gg D_{\text{eff}}\)
\end{itemize}

We apply a 5--10x safety margin beyond the classical 10--20
samples-per-predictor rule to account for: 1. \emph{Covariate shift}:
OOD testing is harder than IID testing 2. \emph{Model selection
variance}: Hyperparameter tuning consumes effective degrees of freedom
3. \emph{Temporal decay}: Even ``stationary'' relationships weaken over
time

The transition zone (\(100 < N/D_{\text{eff}} < 1000\)) corresponds to
the region where model selection becomes critical and the Compression
Superiority Test is recommended.

\begin{center}\rule{0.5\linewidth}{0.5pt}\end{center}

\subsection{Supplementary Note 7: Epistemic Compression vs.~Classical
Regularization}\label{supplementary-note-7-epistemic-compression-vs.-classical-regularization}

A natural question is how Epistemic Compression differs from standard
regularization techniques such as L1/L2 penalties, dropout, or early
stopping.

\subsubsection{Key Distinctions}\label{key-distinctions}

\begin{longtable}[]{@{}
  >{\raggedright\arraybackslash}p{(\columnwidth - 4\tabcolsep) * \real{0.1455}}
  >{\raggedright\arraybackslash}p{(\columnwidth - 4\tabcolsep) * \real{0.4545}}
  >{\raggedright\arraybackslash}p{(\columnwidth - 4\tabcolsep) * \real{0.4000}}@{}}
\toprule\noalign{}
\begin{minipage}[b]{\linewidth}\raggedright
Aspect
\end{minipage} & \begin{minipage}[b]{\linewidth}\raggedright
Classical Regularization
\end{minipage} & \begin{minipage}[b]{\linewidth}\raggedright
Epistemic Compression
\end{minipage} \\
\midrule\noalign{}
\endhead
\bottomrule\noalign{}
\endlastfoot
\emph{Mechanism} & Post-hoc penalty on weights & Architectural
constraint on representable functions \\
\emph{Flexibility} & Model \emph{can} represent complex functions;
regularization discourages it & Model \emph{cannot} represent certain
functions; architecture forbids it \\
\emph{Analogy} & A leash (can be pulled) & A fence (cannot be
crossed) \\
\emph{Failure Mode} & Can be overridden by strong gradients &
Architecturally resistant by design; empirical validation in new domains
required \\
\emph{Interpretability} & Weights remain dense, hard to interpret &
Sparse/structured, often interpretable \\
\emph{Domain Knowledge} & Not encoded & Explicitly encoded in
architecture \\
\end{longtable}

\subsubsection{Examples}\label{examples}

\emph{Regularization (Leash):} A neural network with L2 penalty can
still fit arbitrary functions if the gradient is strong enough. The
penalty \emph{discourages} overfitting but does not \emph{prevent} it.

\emph{Epistemic Compression (Fence):} - A \emph{pharmacokinetic
compartment model} cannot learn that ``blood draw timing'' predicts drug
response because mass-action kinetics are hard-coded---there is no
parameter for this. - A \emph{CRATE layer} cannot memorize
high-frequency noise because its compression objective geometrically
prevents it---the bit-cost exceeds the information gain. - A
\emph{3-variable clinical score} cannot overfit to EHR artifacts because
it simply lacks the capacity to represent them.

\subsubsection{Implication}\label{implication}

Regularization is appropriate when you trust the data and want to smooth
the learned function. Epistemic Compression is appropriate when you
\emph{distrust} the data and want to structurally prevent the model from
learning patterns that cannot be trusted.

\begin{center}\rule{0.5\linewidth}{0.5pt}\end{center}

\section{References}\label{references}

\phantomsection\label{refs}


\begin{CSLReferences}{0}{1}
\bibitem[\citeproctext]{ref-jumper2021}
\CSLLeftMargin{1. }%
\CSLRightInline{Jumper J, Evans R, Pritzel A, et al. Highly accurate
protein structure prediction with {AlphaFold}. Nature.
2021;596(7873):583--9.
doi:\href{https://doi.org/10.1038/s41586-021-03819-2}{10.1038/s41586-021-03819-2}}

\bibitem[\citeproctext]{ref-kapoor2023}
\CSLLeftMargin{2. }%
\CSLRightInline{Kapoor S, Narayanan A. Leakage and the reproducibility
crisis in machine-learning-based science. Patterns. 2023;4(9).
doi:\href{https://doi.org/10.1016/j.patter.2023.100804}{10.1016/j.patter.2023.100804}}

\bibitem[\citeproctext]{ref-nestor2019feature}
\CSLLeftMargin{3. }%
\CSLRightInline{Nestor CC, Haibe-Kains B, Goldenberg A, et al. Feature
robustness in non-stationary health records: Caveats to deployable
clinical machine learning. In: Proceedings of the 4th machine learning
for healthcare conference (MLHC) {[}Internet{]}. 2019. p. 1--23.
Available from: \url{https://proceedings.mlr.press/v106/nestor19a.html}}

\bibitem[\citeproctext]{ref-breiman2001}
\CSLLeftMargin{4. }%
\CSLRightInline{Breiman L. Statistical modeling: The two cultures.
Statistical Science. 2001;16(3):199--231. }

\bibitem[\citeproctext]{ref-geman1992}
\CSLLeftMargin{5. }%
\CSLRightInline{Geman S, Bienenstock E, Doursat R. Neural networks and
the bias/variance dilemma. Neural Computation. 1992;4(1):1--58.
doi:\href{https://doi.org/10.1162/neco.1992.4.1.1}{10.1162/neco.1992.4.1.1}}

\bibitem[\citeproctext]{ref-demiguel2009}
\CSLLeftMargin{6. }%
\CSLRightInline{DeMiguel V, Garlappi L, Uppal R. Optimal versus naive
diversification: How inefficient is the {1/N} portfolio strategy? Review
of Financial Studies. 2009;22(5):1915--53.
doi:\href{https://doi.org/10.1093/rfs/hhm075}{10.1093/rfs/hhm075}}

\bibitem[\citeproctext]{ref-wong2021sepsis}
\CSLLeftMargin{7. }%
\CSLRightInline{Wong A, Otles E, Donnelly JP, Krumm A, McCullough JT,
Deis O, et al. External validation of a widely implemented sepsis
prediction model in hospitalized patients. JAMA Internal Medicine
{[}Internet{]}. 2021;181(8):1065--70.
doi:\href{https://doi.org/10.1001/jamainternmed.2021.2626}{10.1001/jamainternmed.2021.2626}}

\bibitem[\citeproctext]{ref-finlayson2021clinician}
\CSLLeftMargin{8. }%
\CSLRightInline{Finlayson SG, Subbaswamy A, Singh K, Bowers J, Kupke A,
Zittrain J, et al. The clinician and dataset shift in artificial
intelligence. New England Journal of Medicine. 2021;385(3):283--6.
doi:\href{https://doi.org/10.1056/NEJMc2104626}{10.1056/NEJMc2104626}}

\bibitem[\citeproctext]{ref-roberts2021}
\CSLLeftMargin{9. }%
\CSLRightInline{Roberts M, Driggs D, Thorpe M, et al. Common pitfalls
and recommendations for using machine learning to detect and
prognosticate for {COVID-19} using chest radiographs and {CT} scans.
Nature Machine Intelligence. 2021;3:199--217.
doi:\href{https://doi.org/10.1038/s42256-021-00307-0}{10.1038/s42256-021-00307-0}}

\bibitem[\citeproctext]{ref-bertelsmann2025fragile}
\CSLLeftMargin{10. }%
\CSLRightInline{Bertelsmann Stiftung. Fragile foundations: Hidden risks
of generative AI {[}Internet{]}. G{ü}tersloh, Germany: Bertelsmann
Stiftung; 2025. Available from:
\url{https://www.bertelsmann-stiftung.de/fileadmin/files/user_upload/Fragile_foundations_risks_of_generativ_AI_2025.pdf}}

\bibitem[\citeproctext]{ref-wynants2020prediction}
\CSLLeftMargin{11. }%
\CSLRightInline{Wynants L, Van Calster B, Collins GS, et al. Prediction
models for diagnosis and prognosis of covid-19: Systematic review and
critical appraisal. BMJ. 2020;369.
doi:\href{https://doi.org/10.1136/bmj.m1328}{10.1136/bmj.m1328}}

\bibitem[\citeproctext]{ref-degrave2021}
\CSLLeftMargin{12. }%
\CSLRightInline{DeGrave AJ, Janizek JD, Lee SI. {AI} for radiographic
{COVID-19} detection selects shortcuts over signal. Nature Machine
Intelligence. 2021;3:610--9.
doi:\href{https://doi.org/10.1038/s42256-021-00338-7}{10.1038/s42256-021-00338-7}}

\bibitem[\citeproctext]{ref-korzybski1933}
\CSLLeftMargin{13. }%
\CSLRightInline{Korzybski A. Science and sanity: An introduction to
non-aristotelian systems and general semantics. Lancaster, PA:
International Non-Aristotelian Library Publishing Company; 1933. }

\bibitem[\citeproctext]{ref-abbas2021}
\CSLLeftMargin{14. }%
\CSLRightInline{Abbas A, Sutter D, Figalli A, Woerner S. Effective
dimension of machine learning models {[}Internet{]}. 2021. Available
from: \url{https://arxiv.org/abs/2112.04807}}

\bibitem[\citeproctext]{ref-nakkiran2019doubledescent}
\CSLLeftMargin{15. }%
\CSLRightInline{Nakkiran P, Kaplun G, Bansal Y, Yang T, Barak B,
Sutskever I. Deep double descent: Where bigger models and more data
hurt. Journal of Statistical Mechanics: Theory and Experiment
{[}Internet{]}. 2021; Available from:
\url{https://arxiv.org/abs/1912.02292}}

\bibitem[\citeproctext]{ref-pearl2009}
\CSLLeftMargin{16. }%
\CSLRightInline{Pearl J. Causality: Models, reasoning, and inference.
2nd ed. Cambridge University Press; 2009. }

\bibitem[\citeproctext]{ref-yu2022ratereduction}
\CSLLeftMargin{17. }%
\CSLRightInline{Yu Y, Chan KHR, You C, Song C, Ma Y. Learning diverse
and discriminative representations via the principle of maximal coding
rate reduction. IEEE Transactions on Pattern Analysis and Machine
Intelligence. 2022;44(12):9144--63.
doi:\href{https://doi.org/10.1109/TPAMI.2021.3124113}{10.1109/TPAMI.2021.3124113}}

\bibitem[\citeproctext]{ref-ma2022parsimony}
\CSLLeftMargin{18. }%
\CSLRightInline{Ma Y, Tsao D, Shum HY. On the principles of parsimony
and self-consistency for the emergence of intelligence. Frontiers of
Information Technology \& Electronic Engineering. 2022;23(9):1298--323.
doi:\href{https://doi.org/10.1631/FITEE.2200297}{10.1631/FITEE.2200297}}

\bibitem[\citeproctext]{ref-tishby2000}
\CSLLeftMargin{19. }%
\CSLRightInline{Tishby N, Pereira FC, Bialek W. The information
bottleneck method. arXiv preprint physics/0004057. 2000; }

\bibitem[\citeproctext]{ref-tishby2015deep}
\CSLLeftMargin{20. }%
\CSLRightInline{Tishby N, Zaslavsky N. Deep learning and the information
bottleneck principle. arXiv preprint arXiv:150302406 {[}Internet{]}.
2015; Available from: \url{https://arxiv.org/abs/1503.02406}}

\bibitem[\citeproctext]{ref-vapnik1998}
\CSLLeftMargin{21. }%
\CSLRightInline{Vapnik VN. Statistical learning theory. Wiley; 1998. }

\bibitem[\citeproctext]{ref-gigerenzer1999}
\CSLLeftMargin{22. }%
\CSLRightInline{Gigerenzer G, Todd PM. Simple heuristics that make us
smart. Oxford University Press; 1999. }

\bibitem[\citeproctext]{ref-simon1956}
\CSLLeftMargin{23. }%
\CSLRightInline{Simon HA. Rational choice and the structure of the
environment. Psychological Review. 1956;63(2):129--38.
doi:\href{https://doi.org/10.1037/h0042769}{10.1037/h0042769}}

\bibitem[\citeproctext]{ref-kahneman2011}
\CSLLeftMargin{24. }%
\CSLRightInline{Kahneman D. Thinking, fast and slow. New York: Farrar,
Straus; Giroux; 2011. }

\bibitem[\citeproctext]{ref-bengio2019system2}
\CSLLeftMargin{25. }%
\CSLRightInline{Bengio Y, Deleu T, Rahaman N, Ke R, Lachapelle S,
Bilaniuk O, et al. A meta-transfer objective for learning to disentangle
causal mechanisms. arXiv preprint arXiv:190110912. 2019; }

\bibitem[\citeproctext]{ref-kuang2020stable}
\CSLLeftMargin{26. }%
\CSLRightInline{Kuang K, Xiong R, Cui P, Athey S, Li B. Stable
prediction with model misspecification and agnostic distribution shift.
In: Proceedings of the AAAI conference on artificial intelligence
{[}Internet{]}. 2020. p. 4485--92.
doi:\href{https://doi.org/10.1609/aaai.v34i04.5876}{10.1609/aaai.v34i04.5876}}

\bibitem[\citeproctext]{ref-sambasivan2021}
\CSLLeftMargin{27. }%
\CSLRightInline{Sambasivan N, Kapania S, Highlander H, Akrong E,
Paritosh P, Goel D. {``Everyone wants to do the model work, not the data
work''}: Data cascades in high-stakes {AI}. In: Proceedings of the 2021
CHI conference on human factors in computing systems. 2021.
doi:\href{https://doi.org/10.1145/3411764.3445617}{10.1145/3411764.3445617}}

\bibitem[\citeproctext]{ref-damour2022underspecification}
\CSLLeftMargin{28. }%
\CSLRightInline{D'Amour A, Heller K, Moldovan D, Adlam B, Alipanahi B,
et al. Underspecification presents challenges for credibility in modern
machine learning. Journal of Machine Learning Research {[}Internet{]}.
2022;23(226):1--61. Available from:
\url{https://www.jmlr.org/papers/v23/20-1335.html}}

\bibitem[\citeproctext]{ref-silver2018alphazero}
\CSLLeftMargin{29. }%
\CSLRightInline{Silver D, Hubert T, Schrittwieser J, Antonoglou I,
Dhariwal M, Jumper J, et al. A general reinforcement learning algorithm
that masters chess, shogi, and go through self-play. Science.
2018;362(6419):1140--4.
doi:\href{https://doi.org/10.1126/science.aar6404}{10.1126/science.aar6404}}

\bibitem[\citeproctext]{ref-pai2024}
\CSLLeftMargin{30. }%
\CSLRightInline{Pai S, Hesse N, et al. Foundation model for cancer
imaging biomarkers. Nature Machine Intelligence. 2024;6(3):354--67.
doi:\href{https://doi.org/10.1038/s42256-024-00807-9}{10.1038/s42256-024-00807-9}}

\bibitem[\citeproctext]{ref-hastie2009}
\CSLLeftMargin{31. }%
\CSLRightInline{Hastie T, Tibshirani R, Friedman J. The elements of
statistical learning: Data mining, inference, and prediction. 2nd ed.
Springer; 2009.
doi:\href{https://doi.org/10.1007/978-0-387-84858-7}{10.1007/978-0-387-84858-7}}

\bibitem[\citeproctext]{ref-harrell2015}
\CSLLeftMargin{32. }%
\CSLRightInline{Harrell FE. Regression modeling strategies: With
applications to linear models, logistic and ordinal regression, and
survival analysis. 2nd ed. Springer; 2015.
doi:\href{https://doi.org/10.1007/978-3-319-19425-7}{10.1007/978-3-319-19425-7}}

\bibitem[\citeproctext]{ref-wolpert1997}
\CSLLeftMargin{33. }%
\CSLRightInline{Wolpert DH, Macready WG. No free lunch theorems for
optimization. IEEE Transactions on Evolutionary Computation.
1997;1(1):67--82.
doi:\href{https://doi.org/10.1109/4235.585893}{10.1109/4235.585893}}

\bibitem[\citeproctext]{ref-gorishniy2021fttransformer}
\CSLLeftMargin{34. }%
\CSLRightInline{Gorishniy Y, Rubachev I, Khrulkov V, Babenko A.
Revisiting deep learning models for tabular data. In: Advances in neural
information processing systems. 2021. p. 18932--43. }

\bibitem[\citeproctext]{ref-becker1996adult}
\CSLLeftMargin{35. }%
\CSLRightInline{Becker B, Kohavi R. Adult {[}Internet{]}. UCI Machine
Learning Repository; 1996.
doi:\href{https://doi.org/10.24432/C5XW20}{10.24432/C5XW20}}

\bibitem[\citeproctext]{ref-shumailov2024collapse}
\CSLLeftMargin{36. }%
\CSLRightInline{Shumailov I, Shumaylov Z, Zhao Y, Gal Y, Papernot N,
Anderson R. AI models collapse when trained on recursively generated
data. Nature. 2024;631:755--9. }

\bibitem[\citeproctext]{ref-schoelkopf2021}
\CSLLeftMargin{37. }%
\CSLRightInline{Schölkopf B, Locatello F, Bauer S, Ke NR, Kalchbrenner
N, Goyal A, et al. Toward causal representation learning. Proceedings of
the IEEE. 2021;109(5):612--34. }

\bibitem[\citeproctext]{ref-arjovsky2020}
\CSLLeftMargin{38. }%
\CSLRightInline{Arjovsky M, Bottou L, Gulrajani I, Lopez-Paz D.
Invariant risk minimization. arXiv preprint arXiv:190702893. 2020; }

\bibitem[\citeproctext]{ref-leeb2025causality}
\CSLLeftMargin{39. }%
\CSLRightInline{Leeb F, Jin Z, Schölkopf B. Causality can systematically
address the monsters under the bench(marks) {[}Internet{]}. 2025.
Available from: \url{https://arxiv.org/abs/2502.05085}}

\bibitem[\citeproctext]{ref-lourie2025scaling}
\CSLLeftMargin{40. }%
\CSLRightInline{Lourie N et al. Scaling laws are unreliable for
downstream tasks: A reality check. In: Findings of the association for
computational linguistics: EMNLP 2025 {[}Internet{]}. Association for
Computational Linguistics; 2025. p. 16167--80. Available from:
\url{https://aclanthology.org/2025.findings-emnlp.877/}}

\bibitem[\citeproctext]{ref-gulrajani2023}
\CSLLeftMargin{41. }%
\CSLRightInline{Gulrajani I, Lopez-Paz D. In search of lost domain
generalization. arXiv preprint arXiv:200701434. 2023; }

\bibitem[\citeproctext]{ref-rothenhausler2021}
\CSLLeftMargin{42. }%
\CSLRightInline{Rothenhaüsler D, Meinshausen N, Bühlmann P, Peters J.
Anchor regression: Heterogeneous data meet causality. Journal of the
Royal Statistical Society: Series B (Statistical Methodology).
2021;83(2):215--46.
doi:\href{https://doi.org/10.1111/rssb.12398}{10.1111/rssb.12398}}

\bibitem[\citeproctext]{ref-floridi2025conjecture}
\CSLLeftMargin{43. }%
\CSLRightInline{Floridi L. A conjecture on a fundamental trade-off
between certainty and scope in symbolic and generative AI. arXiv
preprint arXiv:250610130 {[}Internet{]}. 2025; Available from:
\url{https://arxiv.org/abs/2506.10130}}

\bibitem[\citeproctext]{ref-spivack2025}
\CSLLeftMargin{44. }%
\CSLRightInline{Spivack N, Jonk G. Epistemology and metacognition in
artificial intelligence: Defining, classifying, and governing the limits
of AI knowledge {[}Internet{]}. Whitepaper; 2025. Available from:
\url{https://www.novaspivack.com/technology/ai-technology/epistemology-and-metacognition-in-artificial-intelligence-defining-classifying-and-governing-the-limits-of-ai-knowledge}}

\bibitem[\citeproctext]{ref-dittrich2025}
\CSLLeftMargin{45. }%
\CSLRightInline{Dittrich C, Kinne JF. The information-theoretic
imperative: Compression and the epistemic foundations of intelligence.
arXiv preprint arXiv:251025883 {[}Internet{]}. 2025; Available from:
\url{https://arxiv.org/abs/2510.25883}}

\bibitem[\citeproctext]{ref-power2022}
\CSLLeftMargin{46. }%
\CSLRightInline{Power A, Burda Y, Edwards H, Babuschkin I, Misra V.
Grokking: Generalization beyond overfitting on small algorithmic
datasets. arXiv preprint arXiv:220102177. 2022; }

\bibitem[\citeproctext]{ref-sutton2019}
\CSLLeftMargin{47. }%
\CSLRightInline{Sutton R. The bitter lesson {[}Internet{]}. Incomplete
Ideas (blog); 2019. Available from:
\url{http://www.incompleteideas.net/IncIdeas/BitterLesson.html}}

\bibitem[\citeproctext]{ref-poggio2020}
\CSLLeftMargin{48. }%
\CSLRightInline{Poggio T, Banburski A, Liao Q. Theoretical issues in
deep networks. Proceedings of the National Academy of Sciences.
2020;117(48):30039--45.
doi:\href{https://doi.org/10.1073/pnas.1907369117}{10.1073/pnas.1907369117}}

\bibitem[\citeproctext]{ref-lake2017building}
\CSLLeftMargin{49. }%
\CSLRightInline{Lake BM, Ullman TD, Tenenbaum JB, Gershman SJ. Building
machines that learn and think like people. Behavioral and Brain
Sciences. 2017;40.
doi:\href{https://doi.org/10.1017/S0140525X16001837}{10.1017/S0140525X16001837}}

\bibitem[\citeproctext]{ref-peirce1923}
\CSLLeftMargin{50. }%
\CSLRightInline{Peirce CS. Chance, love, and logic: Philosophical
essays. Harcourt, Brace \& Company; 1923. }

\bibitem[\citeproctext]{ref-polanyi1962}
\CSLLeftMargin{51. }%
\CSLRightInline{Polanyi M. Personal knowledge: Towards a post-critical
philosophy. University of Chicago Press; 1962. }

\bibitem[\citeproctext]{ref-wornow2023shaky}
\CSLLeftMargin{52. }%
\CSLRightInline{Wornow M, Xu Y, Thapa R, Patel B, Steinberg E, Fleming
S, et al. The shaky foundations of clinical foundation models: A survey
of large language models and foundation models for {EMRs}. npj Digital
Medicine. 2023;6(1):135.
doi:\href{https://doi.org/10.1038/s41746-023-00879-8}{10.1038/s41746-023-00879-8}}

\bibitem[\citeproctext]{ref-geirhos2020}
\CSLLeftMargin{53. }%
\CSLRightInline{Geirhos R, Jacobsen JH, Michaelis C, et al. Shortcut
learning in deep neural networks. Nature Machine Intelligence.
2020;2(11):665--73.
doi:\href{https://doi.org/10.1038/s42256-020-00257-z}{10.1038/s42256-020-00257-z}}

\bibitem[\citeproctext]{ref-mitchell2023}
\CSLLeftMargin{54. }%
\CSLRightInline{Mitchell M, Krakauer DC. The debate over understanding
in {AI}'s large language models. Proceedings of the National Academy of
Sciences {[}Internet{]}. 2023 Mar;120(13).
doi:\href{https://doi.org/10.1073/pnas.2215907120}{10.1073/pnas.2215907120}}

\bibitem[\citeproctext]{ref-kim2025}
\CSLLeftMargin{55. }%
\CSLRightInline{Kim Y, Jeong H, Chen S, Li SS, Lu M, Alhamoud K, et al.
Medical hallucination in foundation models and their impact on
healthcare. medRxiv {[}Internet{]}. 2025;
doi:\href{https://doi.org/10.1101/2025.02.28.25323115}{10.1101/2025.02.28.25323115}}

\bibitem[\citeproctext]{ref-nori2023capabilities}
\CSLLeftMargin{56. }%
\CSLRightInline{Nori H, King N, McKinney SM, et al. Capabilities of
GPT-4 on medical challenge problems. arXiv preprint arXiv:230313375.
2023; }

\bibitem[\citeproctext]{ref-singhal2023large}
\CSLLeftMargin{57. }%
\CSLRightInline{Singhal K, Azizi S, Tu T, et al. Large language models
encode clinical knowledge. Nature. 2023;620:172--80.
doi:\href{https://doi.org/10.1038/s41586-023-06291-2}{10.1038/s41586-023-06291-2}}

\bibitem[\citeproctext]{ref-tizhoosh2025beyond}
\CSLLeftMargin{58. }%
\CSLRightInline{Tizhoosh HR. Beyond the failures: Rethinking foundation
models in pathology {[}Internet{]}. 2025. Available from:
\url{https://arxiv.org/abs/2510.23807}}

\bibitem[\citeproctext]{ref-ouyang2022training}
\CSLLeftMargin{59. }%
\CSLRightInline{Ouyang L, Wu J, Jiang X, Almeida D, Wainwright C,
Mishkin P, et al. Training language models to follow instructions with
human feedback. Advances in Neural Information Processing Systems.
2022;35:27730--44. }

\bibitem[\citeproctext]{ref-hu2022lora}
\CSLLeftMargin{60. }%
\CSLRightInline{Hu EJ, Shen Y, Wallis P, Allen-Zhu Z, Li Y, Wang S, et
al. LoRA: Low-rank adaptation of large language models. In:
International conference on learning representations {[}Internet{]}.
2022. Available from: \url{https://openreview.net/forum?id=nZeVKeeFYf9}}

\bibitem[\citeproctext]{ref-liu2025kan}
\CSLLeftMargin{61. }%
\CSLRightInline{Liu Z, Wang Y, Vaidya S, Ruehle F, Halverson J, Soljačić
M, et al. KAN: Kolmogorov-arnold networks {[}Internet{]}. 2025.
Available from: \url{https://arxiv.org/abs/2404.19756}}

\bibitem[\citeproctext]{ref-hasani2021liquid}
\CSLLeftMargin{62. }%
\CSLRightInline{Hasani R, Lechner M, Amini A, Rus D, Grosu R. Liquid
time-constant networks. Proceedings of the AAAI Conference on Artificial
Intelligence {[}Internet{]}. 2021;35(9):7657--66. Available from:
\url{https://ojs.aaai.org/index.php/AAAI/article/view/16936}}

\bibitem[\citeproctext]{ref-bronstein2021geometric}
\CSLLeftMargin{63. }%
\CSLRightInline{Bronstein MM, Bruna J, Cohen T, Veličković P. {Geometric
Deep Learning: Grids, Groups, Graphs, Geodesics, and Gauges}. arXiv
preprint arXiv:210413478 {[}Internet{]}. 2021; Available from:
\url{https://geometricdeeplearning.com/}}

\bibitem[\citeproctext]{ref-hollmann2025}
\CSLLeftMargin{64. }%
\CSLRightInline{Hollmann N, Müller S, et al. Accurate predictions on
small data with a tabular foundation model. Nature. 2025;637:319--26.
doi:\href{https://doi.org/10.1038/s41586-024-08328-6}{10.1038/s41586-024-08328-6}}

\bibitem[\citeproctext]{ref-yu2024crate}
\CSLLeftMargin{65. }%
\CSLRightInline{Yu Y, Buchanan S, Pai D, Chu T, Wu Z, Tong S, et al.
White-box transformers via sparse rate reduction: Compression is all
there is? Journal of Machine Learning Research. 2024;25(300):1--128. }

\bibitem[\citeproctext]{ref-fanconi2025cascaded}
\CSLLeftMargin{66. }%
\CSLRightInline{Fanconi C, Schaar M van der. Cascaded language models
for cost-effective human-AI decision-making {[}Internet{]}. 2025.
Available from: \url{https://arxiv.org/abs/2506.11887}}

\bibitem[\citeproctext]{ref-ranisch2025foundation}
\CSLLeftMargin{67. }%
\CSLRightInline{Ranisch R, Haltaufderheide J. Foundation models in
medicine are a social experiment: Time for an ethical framework. npj
Digital Medicine. 2025 Aug 16;8(1):525.
doi:\href{https://doi.org/10.1038/s41746-025-01924-4}{10.1038/s41746-025-01924-4}}

\bibitem[\citeproctext]{ref-bommasani2021}
\CSLLeftMargin{68. }%
\CSLRightInline{Bommasani R, Hudson DA, Adeli E, Altman R, Arora S, Arx
S von, et al. On the opportunities and risks of foundation models. arXiv
preprint arXiv:210807258. 2021; }

\bibitem[\citeproctext]{ref-rudin2019}
\CSLLeftMargin{69. }%
\CSLRightInline{Rudin C. Stop explaining black box machine learning
models for high stakes decisions and use interpretable models instead.
Nature Machine Intelligence. 2019;1(5):206--15.
doi:\href{https://doi.org/10.1038/s42256-019-0048-x}{10.1038/s42256-019-0048-x}}

\bibitem[\citeproctext]{ref-dietvorst2015}
\CSLLeftMargin{70. }%
\CSLRightInline{Dietvorst BJ, Simmons JP, Massey C. Algorithm aversion:
People erroneously avoid algorithms after seeing them err. Journal of
Experimental Psychology: General. 2015;144(1):114--26.
doi:\href{https://doi.org/10.1037/xge0000033}{10.1037/xge0000033}}

\bibitem[\citeproctext]{ref-collins2024tripodai}
\CSLLeftMargin{71. }%
\CSLRightInline{Collins GS, Moons KGM, Dhiman P, Riley RD, Beam AL, Van
Calster B, et al. {TRIPOD+AI} statement: Updated guidance for reporting
clinical prediction models that use regression or machine learning
methods. BMJ. 2024;385:e078378.
doi:\href{https://doi.org/10.1136/bmj-2023-078378}{10.1136/bmj-2023-078378}}

\bibitem[\citeproctext]{ref-khutsishvili2025}
\CSLLeftMargin{72. }%
\CSLRightInline{Khutsishvili K. A false confidence in the {EU AI Act}:
Epistemic gaps and bureaucratic traps {[}Internet{]}. Tech Policy Press;
2025. Available from:
\url{https://www.techpolicy.press/a-false-confidence-in-the-eu-ai-act-epistemic-gaps-and-bureaucratic-traps/}}

\bibitem[\citeproctext]{ref-wordsforthewise2019}
\CSLLeftMargin{73. }%
\CSLRightInline{wordsforthewise. {LendingClub} loan data 2007--2018.
Kaggle dataset.
\url{https://www.kaggle.com/datasets/wordsforthewise/lending-club};
2019. }

\bibitem[\citeproctext]{ref-johnson2024mimiciv}
\CSLLeftMargin{74. }%
\CSLRightInline{Johnson A, Bulgarelli L, Pollard T, Gow B, Moody B,
Horng S, et al. MIMIC-IV. PhysioNet {[}Internet{]}. 2024 Oct;
doi:\href{https://doi.org/10.13026/kpb9-mt58}{10.13026/kpb9-mt58}}

\bibitem[\citeproctext]{ref-johnson2023mimiciv}
\CSLLeftMargin{75. }%
\CSLRightInline{Johnson AEW, Bulgarelli L, Shen L, Gayles A, Shammout A,
Horng S, et al. MIMIC-IV, a freely accessible electronic health record
dataset. Scientific Data {[}Internet{]}. 2023;10:1.
doi:\href{https://doi.org/10.1038/s41597-022-01899-x}{10.1038/s41597-022-01899-x}}

\bibitem[\citeproctext]{ref-goldberger2000physiobank}
\CSLLeftMargin{76. }%
\CSLRightInline{Goldberger AL, Amaral LAN, Glass L, Hausdorff JM, Ivanov
PCh, Mark RG, et al. PhysioBank, PhysioToolkit, and PhysioNet:
Components of a new research resource for complex physiologic signals.
Circulation. 2000;101(23):e215--20. }

\bibitem[\citeproctext]{ref-moor2023}
\CSLLeftMargin{77. }%
\CSLRightInline{Moor M, Banerjee O, Abad ZSH, et al. Foundation models
for generalist medical artificial intelligence. Nature.
2023;616:259--65.
doi:\href{https://doi.org/10.1038/s41586-023-05881-4}{10.1038/s41586-023-05881-4}}

\bibitem[\citeproctext]{ref-facco2017}
\CSLLeftMargin{78. }%
\CSLRightInline{Facco E, d'Errico M, Rodriguez A, Laio A. Estimating the
intrinsic dimension of datasets by a minimal neighborhood information.
Scientific Reports. 2017;7(1):12140.
doi:\href{https://doi.org/10.1038/s41598-017-11873-y}{10.1038/s41598-017-11873-y}}

\bibitem[\citeproctext]{ref-sendak2020}
\CSLLeftMargin{79. }%
\CSLRightInline{Sendak MP, Ratliff W, Gao M, et al. Real-world
integration of a sepsis deep learning technology into routine clinical
care: Implementation study. JMIR Medical Informatics. 2020;8(7):e15182.
doi:\href{https://doi.org/10.2196/15182}{10.2196/15182}}

\bibitem[\citeproctext]{ref-vanwalraven2010}
\CSLLeftMargin{80. }%
\CSLRightInline{Walraven C van, Dhalla IA, Bell C, Etchells E, Stiell
IG, Zarnke K, et al. Derivation and validation of an index to predict
early death or unplanned readmission after discharge from hospital to
the community. CMAJ. 2010;182(15):1637--44.
doi:\href{https://doi.org/10.1503/cmaj.091974}{10.1503/cmaj.091974}}

\bibitem[\citeproctext]{ref-sparano2018}
\CSLLeftMargin{81. }%
\CSLRightInline{Sparano JA, Gray RJ, Makower DF, et al. Adjuvant
chemotherapy guided by a 21-gene expression assay in breast cancer. New
England Journal of Medicine. 2018;379(2):111--21.
doi:\href{https://doi.org/10.1056/NEJMoa1804710}{10.1056/NEJMoa1804710}}

\bibitem[\citeproctext]{ref-makridakis2018}
\CSLLeftMargin{82. }%
\CSLRightInline{Makridakis S, Spiliotis E, Assimakopoulos V. Statistical
and machine learning forecasting methods: Concerns and ways forward.
PLoS ONE. 2018;13(3):e0194889.
doi:\href{https://doi.org/10.1371/journal.pone.0194889}{10.1371/journal.pone.0194889}}

\bibitem[\citeproctext]{ref-lam2023graphcast}
\CSLLeftMargin{83. }%
\CSLRightInline{Lam R, Sanchez-Gonzalez A, Willson M, Wirnsberger P,
Rasp M, Fortunato M, et al. GraphCast: Learning skillful medium-range
global weather forecasting. Science. 2023;382(6677):1416--21.
doi:\href{https://doi.org/10.1126/science.adi2336}{10.1126/science.adi2336}}

\bibitem[\citeproctext]{ref-russakovsky2015}
\CSLLeftMargin{84. }%
\CSLRightInline{Russakovsky O, Deng J, Su H, Krause J, Satheesh S, Ma S,
et al. ImageNet large scale visual recognition challenge. International
Journal of Computer Vision. 2015;115(3):211--52.
doi:\href{https://doi.org/10.1007/s11263-015-0816-y}{10.1007/s11263-015-0816-y}}

\bibitem[\citeproctext]{ref-vaswani2017}
\CSLLeftMargin{85. }%
\CSLRightInline{Vaswani A, Shazeer N, Parmar N, Uszkoreit J, Jones L,
Gomez AN, et al. Attention is all you need. In: Advances in neural
information processing systems {[}Internet{]}. 2017. Available from:
\url{https://proceedings.neurips.cc/paper_files/paper/2017/file/3f5ee243547dee91fbd053c1c4a845aa-Paper.pdf}}

\bibitem[\citeproctext]{ref-frankle2019}
\CSLLeftMargin{86. }%
\CSLRightInline{Frankle J, Carlin M. The lottery ticket hypothesis:
Finding sparse, trainable neural networks. arXiv preprint
arXiv:180303635. 2019; }

\end{CSLReferences}
\end{document}